\documentclass[journal]{IEEEtran}
\usepackage{amsfonts}
\usepackage{algorithmic}
\usepackage{algorithm}
\usepackage{array}
\usepackage[caption=false,font=normalsize,labelfont=sf,textfont=sf]{subfig}
\usepackage{textcomp}
\usepackage{stfloats}
\usepackage{url}
\usepackage{verbatim}
\usepackage{graphicx}
\usepackage{cite}
\usepackage{bm}
\usepackage{float}
\usepackage{hyperref}
\usepackage{amsmath,amssymb,booktabs}
\usepackage{multirow}
\usepackage{bookmark}
\usepackage{color}
\usepackage{etoolbox}
\makeatletter
\patchcmd{\@makecaption}
  {\scshape}
  {}
  {}
  {}
\makeatother
\begin{document}

\title{SSFam: Scribble Supervised Salient Object Detection Family}

\author{Zhengyi Liu, Sheng Deng, Xinrui Wang, Linbo Wang, Xianyong Fang, Bin Tang*
\thanks{This work is supported by the Natural Science Foundation of Anhui Province (2108085MF210), Key Program of Natural Science
Project of Educational Commission of Anhui Province (KJ2021A0042), and the Talent Research Fund Project of Hefei University (21-22RC14)  (Corresponding author: Bin Tang)}
\thanks{Zhengyi Liu, Sheng Deng, Xinrui Wang, Linbo Wang, and Xianyong Fang are with Key Laboratory of Intelligent Computing and Signal Processing of Ministry of Education, School of Computer Science and Technology, Anhui University, Hefei, China(e-mail: liuzywen@ahu.edu.cn, 2647516703@qq.com, 2325687760@qq.com, wanglb@ahu.edu.cn, fangxianyong@ahu.edu.cn)}
\thanks{Bin Tang is with School of Artificial Intelligence and Big Data, Hefei University, Hefei, China(e-mail: 424539820@qq.com).}
}

\markboth{Journal of \LaTeX\ Class Files,~Vol.~14, No.~8, January~2024}%
{Shell \MakeLowercase{\textit{et al.}}: A Sample Article Using IEEEtran.cls for IEEE Journals}

\IEEEpubid{0000--0000/00\$00.00~\copyright~2024 IEEE}

\maketitle

\begin{abstract}
Scribble supervised salient object detection (SSSOD) constructs segmentation ability of attractive objects from surroundings under the supervision of sparse scribble labels. For the better segmentation, depth and thermal infrared modalities serve as the supplement to RGB images in the complex scenes. Existing methods specifically design various feature extraction and multi-modal fusion strategies for RGB, RGB-Depth, RGB-Thermal, and Visual-Depth-Thermal image input respectively, leading to similar model flood. As the recently proposed Segment Anything Model (SAM) possesses extraordinary segmentation and prompt interactive capability, we propose an SSSOD family based on SAM, named \textit{SSFam}, for the combination input with different modalities. Firstly, different modal-aware modulators are designed to attain modal-specific knowledge which cooperates with modal-agnostic information extracted from the frozen SAM encoder for the better feature ensemble. Secondly, a siamese decoder is tailored to bridge the gap between the training with scribble prompt and the testing with no prompt for the stronger decoding ability. Our model demonstrates the remarkable performance among combinations of different modalities and refreshes the highest level of scribble supervised methods and comes close to the ones of fully supervised methods. \href{https://github.com/liuzywen/SSFam}{https://github.com/liuzywen/SSFam}
\end{abstract}


\section{Introduction}
Salient object detection (SOD) focuses on the identification and segmentation of  most visually prominent objects in RGB images, which benefits the other computer vision tasks, e.g., semantic segmentation \cite{chen2022saliency,zhou2020sal,lee2021railroad}, 
object tracking \cite{zhou2021saliency,liu2023tracking}, person and vehicle re-identification \cite{chen2020salience,ren2022s,qian2022navigating}, and audio-visual quality assessment \cite{cao2023attention}.

The prevalent fully supervised methods design and train deep learning model with high-cost pixel-level annotations,
while the Scribble Supervised Salient Object Detection (SSSOD) methods \cite{zhang2020weakly} use the pair of input image and sparse scribble labels indicating foreground and background to train the model and use input image for testing, showing the potential due to less human intervention and moderate supervision.

\begin{figure}[!htp]
\center
  \includegraphics[width=1\linewidth]{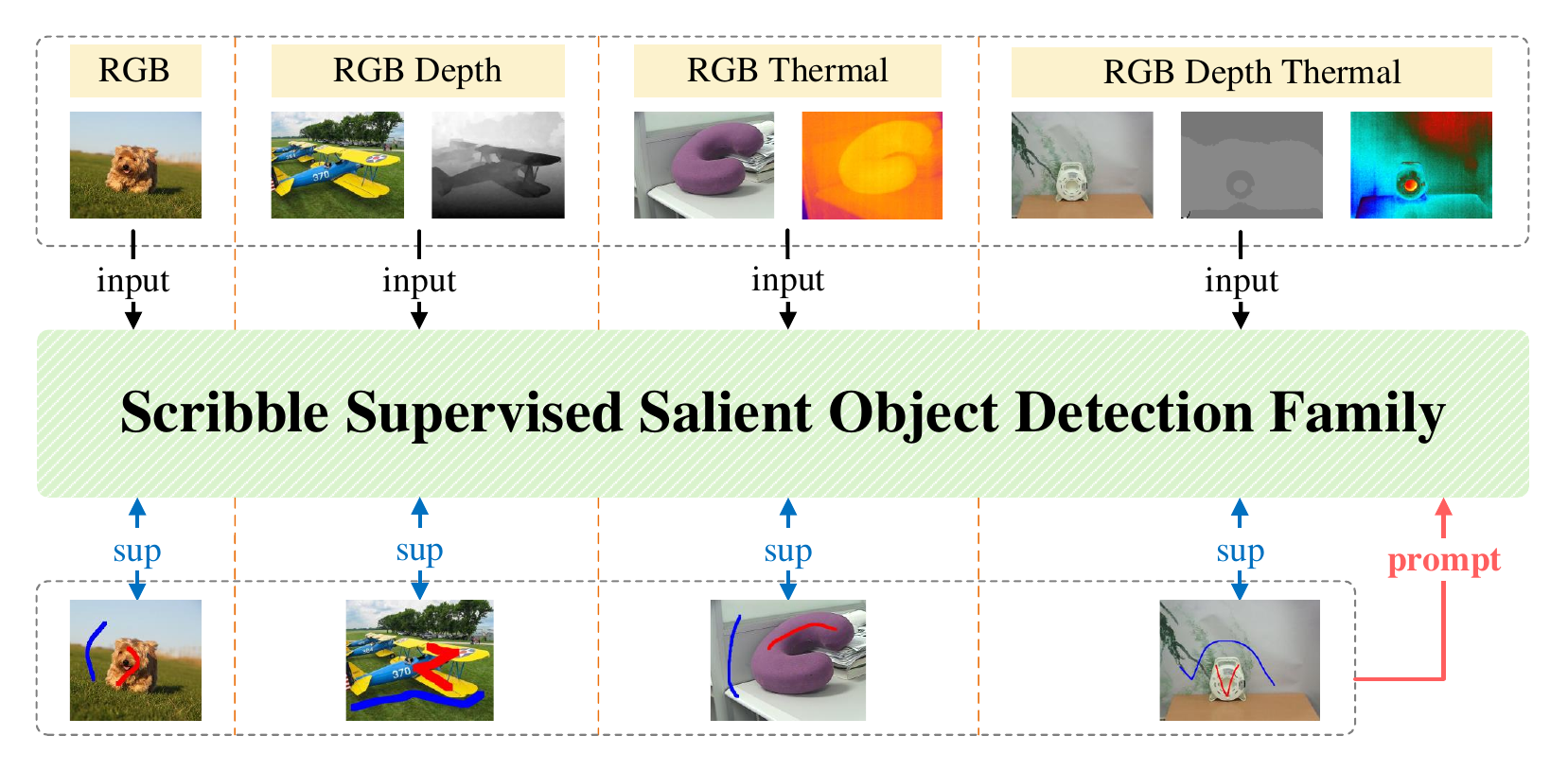}
  \caption{{Scribble supervised salient object detection family for unimodal, bimodal, and trimodal images. }
  \label{fig:Abstract}}
\end{figure}

Existing SSSOD methods \cite{yu2021structure,xu2023synthesize,xu2023visual} borrow pretrained CNN and Transformer encoders which are originally designed for classification task, and then cooperate with progressively upsampling decoders.
The performance is restrained due to the gap between classification task and segmentation one.

\IEEEpubidadjcol
The recent proposed large model Segment Anything Model (SAM) \cite{kirillov2023segment} pretrained for segmentation tasks is naturally a better choice to solve SSSOD task.
Some pioneering researches \cite{chen2023sam,chen2023make,yan2023ringmo,zhang2023customized,ge2023fine,liu2023explicit,cui2023adaptive,li2023refsam,lin2023samus} have fine-tuned SAM to specific unimodal  tasks via parameter-efficient transfer learning techniques \cite{houlsby2019parameter}.

However, in real-world scenes, RGB cameras, depth and thermal infrared sensors are  combinational equipped  to improve visual perception ability.  Accordingly, bimodal SOD, such as RGB-D \cite{sun2023catnet,fu2021siamese,yao2023depth,niu2020boundary} and RGB-T \cite{cong2022does}, and trimodal visible-depth-thermal (V-D-T)  SOD \cite{song2022novel} are tailored to suit the interaction requirements of different modalities, which increase modeling overhead.

Facing the  input challenge with combinations of different modalities, we propose an SSSOD family for unimodal, bimodal, and trimodal input, named \textit{SSFam} (Fig \ref{fig:Abstract}).
Some  modal-aware modulators are attached to SAM and responsible for obtaining modal-specific features,  which will be aggregated into the modal-agnostic features from the frozen SAM encoder by element-wise scaled sum for a better feature ensemble.

Besides, the large model SAM supports  point,  box, and mask prompt to generate the class-agnostic mask by indicating  object position. VRP-SAM \cite{sun2024vrp} designs a visual reference prompt to extend the visual reference segmentation capabilities of SAM.
The scribbles in the SSSOD task supervise the training of the model in the training stage, and meanwhile they can be regarded as prompt of SAM.
Concretely, some points are randomly extracted from the scribbles and then fed into  the prompt encoder of SAM, named scribble prompt, which  extending the weakly supervised learning capabilities of SAM.
However, there is no prompt in the testing stage.
To eliminate the gap, a siamese decoder  is proposed to transfer the parameters with prompts in the training to the ones with no prompt in the testing.


In summary, the main contributions are summarized as follows:
\begin{itemize}
  \item To the best of our knowledge, we are the first to use the large model SAM to solve  scribble supervised multi-modal salient object detection tasks involved RGB, RGB-D, RGB-T, V-D-T combinations of multiple modalities.
  \item To reduce the number of trainable parameters, modal-aware modulators are designed to extract modal-specific features besides modal-agnostic features from the frozen SAM encoder. 
  \item To eliminate the gap between scribble prompt training and no prompt testing, a siamese decoder is proposed to transfer the parameters with prompts in the training to the ones with no prompt in the testing.
  \item Experiments show our approach significantly outperforms the scribble supervised  competitors and partially surpasses the fully supervised ones.
  \item The scribble labels are labeled in V-D-T SOD dataset for the first time, which provide convenience for researches.
\end{itemize}

\section{Related works}

\subsection{Unified framework involved salient object detection}
Salient object detection task focuses on identifying prominent objects from their surroundings. When scenes are complex, the depth and thermal infrared cue from corresponding sensors can provide the supplement to RGB modality. Accordingly, RGB-D, RGB-T, and V-D-T SOD are derived.
For example, DCF \cite{ji2021calibrated} calibrates unreliable raw depth maps first and then fuses  both RGB and depth modalities.
DSU \cite{ji2021promoting} is an unsupervised RGB-D method. The pseudo
labels are optimized via the depth cue. Both methods exploit the roles of depth images on RGB-D SOD task.
Recently, some generalist models are proposed based on the consideration that RGB, RGB-D, and RGB-T SOD possess some commonalities.
AiOSOD \cite{jia2023all} is a unified model for RGB, RGB-D, and RGB-T SOD tasks. RGB-RGB, RGB-D, and RGB-T paired images are concatenated input in batch dimension to guarantee the consistency.
Similarity, VSCode \cite{luo2023vscode} simultaneously trains multi-modal salient object detection and camouflaged object detection tasks.
Some 2D prompts are learnt to perceive the peculiarities on different modalities and tasks. 
UniSOD \cite{wang2023unified} individually trains the modal-aware prompts after pre-train stage for single-modal and multi-modal SOD tasks.
Considering some binary segmentation tasks share vital fundamental similarities,
EVP \cite{liu2023explicit} presents a unified framework for salient object detection, forgery detection, defocus blur detection, shadow detection, and camouflage object detection. The task-specific knowledge is individually learnt by using a few of adapters \cite{chen2022adaptformer}.
An arbitrary modality SOD method
AMSOD \cite{huang2024salient} is proposed for the robustness with  different types and numbers of input modalities.
Besides, some unsupervised methods are also proposed to achieve a unified framework, such as A2S-v2 \cite{zhou2023texture}, A2S-v3 \cite{yuan2024unified}.

Unlike the methods mentioned above, which mainly concentrate on fully supervised methods  and unsupervised ones, our \textit{SSFam} is the first unified model for scribble supervised methods among RGB, RGB-D, RGB-T, V-D-T SOD tasks. At first, modality is ranged from unimodal, bimodal to trimodal. Secondly, the scribbles are regarded as the  prompt to assist segmentation. The designed modal-aware modulators and the siamese decoder are used to address the two challenges. 

\subsection{Weakly supervised salient object detection}
The weakly supervised salient object detection methods use category,  bounding boxes, points, subitizing, hybrid labels, scribbles to supervise the training of the models, which show the  potential of leveraging the  performance and annotation cost.

The category supervised methods \cite{wang2017learning,li2018weakly,piao2021mfnet,piao2022noise,li2021joint} optimize the class activation maps (CAM) which supervise the training of saliency model.
MSW \cite{zeng2019multi} and JSM \cite{li2021joint} further integrate the caption information besides image-level tags.
In JSM\cite{li2021joint}, RGB images are used to predict the saliency maps which are further refined by estimated depth information. Next, initial handcrafted pseudo labels and the refined saliency maps are weighted combined by category predictions to form the updated pseudo labels. Last, a saliency network is trained by the updated pseudo labels.
The box supervised methods \cite{liu2021weakly,liang2022weakly} update pseudo labels guided by bounding boxes.
The point supervised method \cite{gao2022weakly} integrates the edge information to generate pseudo labels due to extremely weak point signals.
The subitizing based method \cite{zheng2021weakly} introduces counting information into SOD.
The method via hybrid labels \cite{cong2022weakly} utilizes a large number of coarse labels generated by the traditional unsupervised method and a small number of real labels.

The scribble based methods use sparse labels indicating foreground and background to achieve the whole object segmentation.
WSSA\cite{zhang2020weakly}, DENet\cite{xu2022weakly}, and RGBTScribble \cite{liu2023RGBTScribble} propose  scribble annotation datasets for RGB, RGB-D, and RGB-T SOD tasks, respectively. The methods explore the structure integrity under the guidance of  edges and superpixels.
Based on WSSA \cite{zhang2020weakly}, SCWSSOD\cite{yu2021structure} design some loss functions to optimize the propagation of sparse annotations and ensure the consistency among different scales.

In our work, we use scribble annotation to train the whole model and test the image without scribble. Due to the strong feature discrimination ability of SAM, our method achieves impressive performance by combining the scribble loss \cite{liu2023RGBTScribble} with our proposed consistency loss which transfers the knowledge with prompt to the ones with no prompt.

\subsection{Improvements involved segment anything model}
SAM has demonstrated its prowess across various fields in terms of unimodal input \cite{ma2023segment,chen2023sam,wu2023medical,zhang2023customized}.
To achieve bimodal application,
ViPT \cite{zhu2023visual} and ProTrack \cite{yang2022prompting} adapt the pre-trained RGB-based foundation model to  multi-modality tracking tasks by learning visual prompt tokens. DUALPATH \cite{park2023dual} transfers the parameters of the image foundation model to  video domain by adapters.
In  audio-visual understanding tasks, a trainable latent audio-visual
hybrid  adapter LAVISH \cite{lin2023vision} is proposed to adapt unimodal ViTs to bimodal domain.

Facing our application with combinations of different modalities  involved unimodal, bimodal, and trimodal, LoRAs with low rank properties are adopted as modal-aware modulators. They are attached to SAM to attain modal-specific knowledge and  meet different input demands.

\section{Method}
\begin{figure*}[!htp]
\center
  \includegraphics[width=1\linewidth]{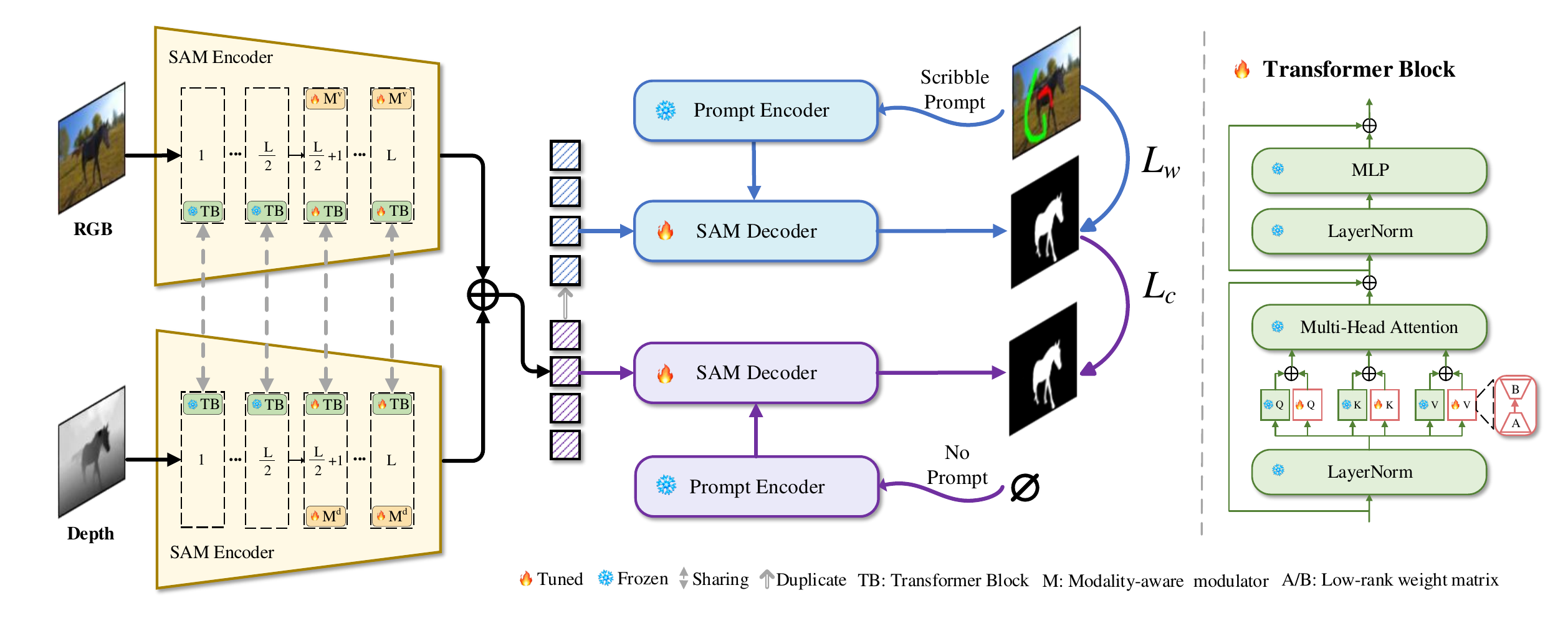}
  \caption{{The proposed scribble supervised salient object detection model. In the  encoding part, a shared and frozen SAM encoder is used to extract modal-agnostic features, and some modal-aware modulators are designed to obtain modal-specific ones. Both features are aggregated by element-wise scaled sum in each block. Last, the features from two modalities are  summed up. In the decoding part, a siamese decoder which consists of a decoder with prompts and a decoder with no prompt is proposed to transfer the parameters with prompts in the training to the ones with no prompt in the testing. When testing, the decoder with no prompt is only needed.}
  \label{fig:Main}}
\end{figure*}
\subsection{SSSOD task description}
To reduce the cost of pixel-level annotations, scribble supervised salient object detection (SSSOD) task takes as the supervision signals the scribbles which are sparse labels indicating foreground and background for training. 
Suppose SSSOD dataset $D=\{(\mathbf{x}_i,\mathbf{y}_i^{{\rm{s}}})\}_{i=1}^{N}$, where $\mathbf{x}_i$ is input image involved unimodal RGB, bimodal RGB-D and RGB-T, and trimodal V-D-T, $\mathbf{y}_i^{\rm{s}}$ is scribble annotation
with 1 representing  foreground, 2 denoting background, and 0 indicating unknown pixels, respectively, and $N$ is the number of training samples. Then, the  SSSOD is to train the model $\mathcal {F} $ with the parameters $ \Omega$ and minimize the loss between predicted saliency maps and scribbles according to the formula:
\begin{equation}
\label{equa:equ1}
    \mathcal{L}(D, \Omega)=\sum_{i=1}^{N}{\mathcal{L}_{w}({\mathcal{F}} _\Omega(\mathbf{x}_i),\mathbf{y}_i^{\rm{s}}))}
\end{equation}
where  the  scribble loss $\mathcal{L}_{w}$  follows  \cite{liu2023RGBTScribble}, and it is defined as:
\begin{equation}
\label{equa:equ2}
    \mathcal{L}_{w}=\mathcal{L}_{pce}+\mathcal{L}_{lsc}+\mathcal{L}_{sl}
\end{equation}
where the partial cross entropy loss ($\mathcal{L}_{pce}$) supervises predicted saliency pixels only corresponding to scribble annotations, the local saliency coherence loss ($\mathcal{L}_{lsc}$) enforces similar pixels in the adjacent region to share consistent saliency scores, and the smoothness loss  ($\mathcal{L}_{sl}$) guarantees the local smoothness and salient distinction along image edges.

\subsection{SSSOD solution}
As a segmentation task, SSSOD can be implemented by Segment Anything Model (SAM). However, due to the lack of domain knowledge, zero-shot performance is poor. Moreover, SAM is pretrained on RGB images instead of bimodal and trimodal images. It is challenging to adapt to input with combinations of different modalities.
To solve the challenge, different modal-aware modulators are designed to attain modal-specific knowledge which cooperates with modal-agnostic information extracted from frozen SAM encoder  for a better feature ensemble.

Besides, the training of SSSOD lies on sparse scribble annotation rather than pixel-level ones.
For a better performance under the less supervision, the model need more prompts. Fortunately, SAM provides the point prompt interface to indicate the position of segmentation.
Some points randomly extracted from scribbles can be regarded as the prompt input of SAM. However, there is no scribble in the testing stage. The gap between the training with prompt and the testing with no prompt becomes the second challenge. To solve the challenge, a siamese decoder is tailored to transfer the knowledge from the decoding branch with prompt to the decoding branch with no prompt.

Accordingly,  we propose an SSSOD family for unimodal, bimodal, and trimodal images, named \textbf{\textit{SSFam}}, by attaching modal-aware modulators to SAM encoder and transforming SAM decoder to a siamese decoder structure. Then, the parameters $\rm \Omega$ of the model $\mathcal{ F}$ is defined as:
\begin{equation}
\label{equa:equ3}
    \Omega=\{\mathrm{\theta}^*_{\rm{E}},\mathrm{\theta}_{\rm{M}}, \mathrm{ \theta}_{\rm{D}}^{\rm{p}},\mathrm{ \theta}_{\rm{D}}^{\rm{n}}\}
\end{equation}
where $\theta_{\rm{E}}^*$ is the parameters of frozen image and prompt encoders in SAM, $\theta_{\rm{M}}$ is the parameters of modulators, $\theta_{\rm{D}}^{\rm{p}}$ is the parameters of the decoder with prompt, and  $\theta_{\rm{D}}^{\rm{n}}$ is the ones of the decoder with no prompt.

\subsection{Two challenges in \textit{SSFam}}
\textbf{Challenge 1 Adapting SAM to input with combinations of different modalities:}
In the real-world scenes, image, depth and thermal infrared sensors are combinational equipped. The three sensors are complementary for detecting salient objects. For example, the color image provides more texture information, depth image offers some spatial geometrical cues, and thermal infrared image supplies scene perception by the temperature differences especially under the dark, rainy, and foggy environments.
As a result, RGB, RGB-D, RGB-T, and V-D-T SSSOD has attracted a lot of research interests. Facing the  input challenge with combinations of different modalities, \textit{SSFam} equips modal-aware modulators  to  attain modal-specific knowledge.
Specifically, $\theta_{\rm{M}}$ is defined as:
\begin{equation}
\label{equa:equ4}
\theta_{\rm{M}}=\left\{
\begin{aligned}
&\{\theta_{\rm{M}}^{\rm{v}}\},&unimodal\\
&\{\theta_{\rm{M}}^{\rm{v}},\theta_{\rm{M}}^{\rm{d}}/\theta_{\rm{M}}^{\rm{t}}\},&bimodal\\
&\{\theta_{\rm{M}}^{\rm{v}},\theta_{\rm{M}}^{\rm{d}},\theta_{\rm{M}}^{\rm{t}}\},&trimodal\\
\end{aligned}
\right.
\end{equation}
where superscript indicates the categories of modality, ${\rm{v}}$, ${\rm{d}}$, ${\rm{t}}$ denotes RGB,  depth, and thermal infrared modality, respectively.

Take as an example RGB-D SSSOD, the model structure of the encoding in \textit{SSFam} is shown in the encoder part of Fig \ref{fig:Main}.
A shared and frozen SAM image encoder is used to extract modal-agnostic features, and two groups of modal-aware modulators are used to learn modal-specific features. The modal-specific feature and the modal-agnostic feature are aggregated by element-wise scaled sum. At last, the features from two modalities are summed up.

The modal-aware modulators can be the Adapters \cite{chen2022adaptformer}, LoRAs \cite{hulora}, SSF \cite{lian2022scaling} and the other fine-tuning solutions \cite{fu2023effectiveness}. Since the number of the parameters in LoRAs are less, LoRAs are chosen as modulators.
Specifically, the input images $\mathbf{x}_i$ are fed into the frozen encoder $\rm{E}$ of SAM. The encoder is a ViT transformer structure with multiple blocks which consist of the multi-head self-attention layers and the feed forward neural layers. Each block ${\rm{E}}_l(l=1,\cdots,L)$ outputs the features $\mathbf{F}_l(l=1,\cdots,L)$. The output of previous block  is fed into the next block.
In the vanilla multi-head self-attention layer,
\begin{equation}
\label{equa:equ5}
\mathbf Q=\mathbf{W}_{\rm{Q}}\mathbf{F}_l,\mathbf{K}=\mathbf{W}_{\rm{K}}\mathbf{F}_l,\mathbf{V}=\mathbf{W}_{\rm{V}}\mathbf{F}_l
\end{equation}
\begin{equation}
\label{equa:equ6}
\mathbf{H}={\rm{Softmax}}(\frac{\mathbf{Q}\mathbf{K}^{\rm{T}}}{\sqrt{\rm{d}}})\mathbf{V}
\end{equation}
where $\mathbf{H}$ is the output of multi-head self-attention layer. It is will fed into feed forward neural layer to generate the output of the transformer block.
\begin{equation}
\label{equa:equ7}
\mathbf{F}_{l+1}={\rm{FFN}}(\mathbf{H})
\end{equation}
where ${\rm{FFN}}(\cdot)$ denotes the feed forward neural layer.

The designed modal-aware modulator introduces low-rank trainable parameters into the multi-head attention layer.
\begin{equation}
\label{equa:equ8}
\mathbf{Q}^m=\mathbf{W}^{m} _{\rm{Q}}{\mathbf{F}}_l,\mathbf{K}^m=\mathbf{W}^{m} _{\rm{K}}{\mathbf{F}}_l,\mathbf{V}^m=\mathbf{W}^{m} _{\rm{V}}{\mathbf{F}}_l
\end{equation}
where $m \in \{\rm v,\rm d\}$ represents modalities in RGB-D SSSOD, and $\mathbf{W}^{m} _n(n \in \{\rm Q,\rm K,\rm V\})$ is composed of the original frozen parameters $\mathbf{W}_n$ and the learnable modal-aware parameters.

\begin{equation}
\left\{
\begin{aligned}
&\mathbf{W}^{m}_{\rm{Q}}=\mathbf{W}_{\rm{Q}}+\mathbf{W}_{\rm{Qa}}^{m}{\mathbf{W}_{\rm{Qb}}^{m}}^{\rm{T}}\\
&\mathbf{W}^{m}_{\rm{K}}=\mathbf{W}_{\rm{K}}+\mathbf{W}_{\rm{Ka}}^{m}{\mathbf{W}_{\rm{Kb}}^{m}}^{\rm{T}}\\
&\mathbf{W}^{m}_{\rm{V}}=\mathbf{W}_{\rm{V}}+\mathbf{W}_{\rm{Va}}^{m}{\mathbf{W}_{\rm{Vb}}^{m}}^{\rm{T}}\\
\end{aligned}
\right.
\end{equation}
where $\mathbf{W}^{m}_{n a}\in \mathbb{R}^{d\times r}$ and $\mathbf{W}^{m}_{n  b}\in \mathbb{R}^{d\times r}$, $r\ll d$. It is achieved by a down-projection linear layer followed by an up-projection linear layer.

At last the features in the last layers of all the modalities  are aggregated by addition operation.
\begin{equation}
\label{equa:equ9}
\mathbf{F}=\sum_{m\in\{v,d\}}{{\mathbf{F}}_L^m}
\end{equation}

Note that it is easily inferred from above discussion about RGB-D example to the others.  $m \in \{{\rm {v}}\}$ in RGB,  $m \in \{{\rm {v}},{\rm {t}}\}$ in RGB-T, and $m \in \{{\rm {v}},{\rm {d}},{\rm {t}}\}$ in V-D-T SSSOD.

\textbf{Challenge 2 Eliminating the gap between training with prompt and testing with no prompt:}
SAM supports point prompt for the better segmentation. The randomly extracted points from scribble labels are fed into SAM in the training stage. However, there is no prompt in the testing.
To eliminate the gap, a siamese decoder is proposed to transfer the parameters with prompts in the training to the ones with no prompt in the testing.  As shown in the decoder part of Fig  \ref{fig:Main},
the feature $\mathbf{F}$ from the encoder is duplicated and fed into prompt decoder branch and no prompt decoder branch, respectively. Then, our \textit{SSFam} is learnt under the loss:

\begin{small}
\begin{equation}
\label{equa:equ10}
    \mathcal{L}(D,\Omega)=\sum_{i=1}^{N}{\mathcal{L}_{w} (\mathcal{F} _{\Omega^{\rm{p}}}(\mathbf{x}_i,\mathbf{y}_i^{\rm{s}}),\mathbf{y}_i^{\rm{s}})+\mathcal{L}_{c}(\mathcal{F}_{\Omega^{\rm{n}}}(\mathbf{x}_i),\mathcal{F}_{\Omega^{\rm{p}}}(\mathbf{x}_i,\mathbf{y}_i^{\rm{s}}))}
\end{equation}
\end{small}

In formula \ref{equa:equ10}, the first loss is scribble supervision loss, which is computed between predicted saliency maps generated from the prompt branch with the parameters $\Omega^{\rm{p}}$=$\{\theta^*_{\rm{E}},\theta_{\rm{M}},\theta^{\rm{p}}_{\rm{D}}\}$ and scribble labels $\mathbf{y}^{\rm{s}}_i$. Note that the input of the  prompt branch include the input image $\mathbf{x}_i$ and scribble prompt $\mathbf{y}^{\rm{s}}_i$. The second loss is a consistency loss $\mathcal{L}_{c}$ which is computed between predicted saliency maps from  the no prompt branch with the parameters $\Omega^{\rm{n}}$=$\{\theta^*_{\rm{E}},\theta_{\rm{M}},\theta^{\rm{n}}_{\rm{D}}\}$
and the ones from the prompt branch with the parameters $\Omega^p$.
The consistency loss  transfers the knowledge with prompt to the ones with no prompt, and it is defined as:
\begin{equation}
\label{equa:equ11}
    \mathcal{L}_{c}=\mathcal{L}_{wbce}+\mathcal{L}_{iou}
\end{equation}
where $\mathcal{L}_{wbce}$ is the weighted binary cross entropy loss and $\mathcal{L}_{iou}$ is weighted IoU loss \cite{wei2020f3net}.

In the end, the training process of \textit{SSFam} is presented in Alg \ref{alg:SSSODAlg}. When testing, the decoder with no prompt is only needed.

\begin{algorithm}[!htp]

	\caption{SSFam training process.}
     \renewcommand{\algorithmicrequire}{\textbf{Input:}}
      \renewcommand{\algorithmicensure}{\textbf{Output:}}
	\label{alg:SSSODAlg}
	\begin{algorithmic}[1]

	\REQUIRE
      The training set  $D=\{(\mathbf{x}_i,\mathbf{y}_i^{\rm{s}})\}_{i=1}^N$, $\mathbf{x}_i$ is RGB, RGB-D, RGB-T, or V-D-T image and $\mathbf{y}_i^{\rm{s}}$ is scribble label; \\

    \ENSURE Model $\mathcal{F}$ with the parameters $\Omega$;\\

 \STATE Load SAM pretrained parameters, including image encoder and prompt encoder $\theta_{\rm{E}}^{*}$,  decoder $\theta_{\rm{D}}^{\rm{p}}$, and another copied decoder $\theta_{\rm{D}}^{\rm{n}}$; \\
 \STATE Initialize modal-specific modulators $\theta_{\rm{M}}^{\rm{v}},\theta_{\rm{M}}^{\rm{d}},\theta_{\rm{M}}^{\rm{t}}$ randomly; \\


\IF{$\mathbf{x}_i$ is RGB}
    \STATE $\theta_{\rm{M}}=\theta_{\rm{M}}^{\rm{v}}$;
\ELSIF{$\mathbf{x}_i$ is RGB-D}
    \STATE $\theta_{\rm{M}}=\{\theta_{\rm{M}}^{\rm{v}},\theta_{\rm{M}}^{\rm{d}}\}$;
\ELSIF{$\mathbf{x}_i$ is RGB-T}
    \STATE $\theta_{\rm{M}}=\{\theta_{\rm{M}}^{\rm{v}},\theta_{\rm{M}}^{\rm{t}}\}$;
\ELSE
    \STATE $\theta_{\rm{M}}=\{\theta_{\rm{M}}^{\rm{v}},\theta_{\rm{M}}^{\rm{d}},\theta_{\rm{M}}^{\rm{t}}\}$;
\ENDIF

\STATE  $\rm \Omega$=$\{\theta_{\rm{E}}^*,\theta_{\rm{D}}^{\rm{p}},\theta_{\rm{D}}^{\rm{n}},\theta_{\rm{M}}\}$;//all the parameters
\STATE  $\rm \Omega^{\rm{p}}$=$\{\theta_{\rm{E}}^*,\theta_{\rm{D}}^{\rm{p}},\theta_{\rm{M}}\}$;
//prompt branch parameters
\STATE  $\Omega^{\rm{n}}$=$\{\theta_{\rm{E}}^*,\theta_{\rm{D}}^{\rm{n}},\theta_{\rm{M}}\}$;//no prompt branch parameters
\STATE  $\Omega^{\rm{pt}}$=$\Omega^{\rm{p}}$-$\theta_{\rm{E}}^*$;//trainable prompt branch  parameters
\STATE  $\Omega^{\rm{nt}}$=$\Omega^{\rm{n}}$-$\theta_{\rm{E}}^*$;//trainable no prompt branch  parameters
\REPEAT
\STATE Take gradient descent step on
\STATE $\bigtriangledown_{\Omega^{\rm{pt}}}\sum_{i=1}^{N}{(\mathcal{L}_{w} ({\mathcal{F} } _{\Omega^{\rm{p}}}(\mathbf{x}_i,\mathbf{y}_i^{\rm{s}}),\mathbf{y}_i^{\rm{s}}))}$\\
$+
\bigtriangledown_{\Omega^{\rm{nt}}}\sum_{i=1}^{N}{(\mathcal{L}_{c}({\mathcal{F}}_{\Omega^{\rm{n}}}(\mathbf{x}_i),F_{\Omega^{\rm{p}}}(\mathbf{x}_i,\mathbf{y}_i^{\rm{s}})))}
$
\UNTIL{converged}

	\RETURN  ${\mathcal{F}}_{\Omega}$.
	\end{algorithmic}
\end{algorithm}

\section{Experiments}
\subsection{Datasets and evaluation metrics}
To verify the effectiveness of \textit{SSFam}, we conduct four groups of comparison experiments involved RGB, RGB-D, RGB-T, and V-D-T scribble supervised  SOD. A summary of datasets is illustrated in Table \ref{tab:Dataset}.

In the unimodal \textbf{RGB} SSSOD comparison experiments, the training dataset adopts 10,553 samples from S-DUTS \cite{zhang2020weakly}.
The testing datasets include
ECSSD\cite{yan2013hierarchical} dataset (1,000 samples), DUTS-Test\cite{wang2017learning} dataset (5,019 samples), HKU-IS\cite{li2015visual} dataset (4,447 samples), DUT-OMRON\cite{yang2013saliency} dataset (5,168 samples), and PASCAL-S\cite{li2014secrets} dataset (850 samples).

In the bimodal \textbf{RGB-D} SSSOD comparison experiments, the training dataset adopts 2,185 paired RGB and depth images from NJU2K-S and NLPR-S \cite{xu2022weakly}.
The testing datasets include
NLPR \cite{peng2014rgbd} with 300 images, NJU2K \cite{ju2014depth} with 500 images, STERE\cite{niu2012leveraging} with 1,000 images, SIP \cite{fan2020rethinking} with 929 images, DES~\cite{cheng2014depth} with 135 images, LFSD\cite{li2014saliency} with 100 images, and SSD \cite{zhu2017three} with 80 images.

In the bimodal  \textbf{RGB-T} SSSOD comparison experiments, the training dataset adopts 2,500 paired RGB and thermal infrared images from RGBT-S \cite{liu2023RGBTScribble}.
The testing datasets include
VT821 \cite{wang2018rgb} with 821 images, VT1000 \cite{tu2019rgb} with 1,000 images, and VT5000 \cite{tu2020rgbt} with 2,500 images.

In the trimodal \textbf{V-D-T} SSSOD comparison experiments, we relabel VDT-2048 SOD \cite{song2022novel} training set via scribbles, which
includes 1,048 paired V-D-T images, namely VDT-S
dataset. The remaining 1,000 samples of VDT-2048 are used to be tested.

\begin{table}[!htp]
\caption{Summary of training and testing datasets.}
\centering
\resizebox{1\linewidth}{!}
{
   \begin{tabular}{c|c|c|c}
\toprule
    Input Type &Dataset Name& $\#$Train & $\#$Test\\

\hline
    \multirow{6}{*}{\centering RGB} &S-DUTS \cite{zhang2020weakly}&10,553&/\\
    &ECSSD\cite{yan2013hierarchical}&/&1,000\\
    &DUTS-Test\cite{wang2017learning}&/&5,019\\
    &HKU-IS\cite{li2015visual}&/&4,447 \\
    &DUT-OMRON\cite{yang2013saliency} &/&5,168\\
    &PASCAL-S\cite{li2014secrets}&/&850\\
\hline
     \multirow{9}{*}{\centering RGB-D}
    &NJU2K-S \cite{xu2022weakly}&1,485 &/\\
    &NLPR-S  \cite{xu2022weakly}&700 &/\\
    &NLPR-Test  \cite{peng2014rgbd}&/&300\\
    &NJU2K-Test  \cite{ju2014depth} &/&500\\
    &STERE\cite{niu2012leveraging} &/&1,000\\
    &SIP \cite{fan2020rethinking} &/&929 \\
    &DES~\cite{cheng2014depth} &/& 135 \\
    &LFSD\cite{li2014saliency}  &/& 100 \\
    &SSD \cite{zhu2017three}  &/& 80 \\
\hline
    \multirow{4}{*}{\centering RGB-T}&RGBT-S \cite{liu2023RGBTScribble}&2,500&/\\
    &VT821 \cite{wang2018rgb}&/& 821\\
    &VT1000 \cite{tu2019rgb}&/& 1,000\\
    &VT5000-Test\cite{tu2020rgbt}&/& 2,500\\
\hline
    \multirow{2}{*}{\centering V-D-T}&VDT-2048-S(\textit{Our proposed})&1,048&/\\
    &VDT-2048-Test\cite{song2022novel}&/&1,000\\
   \bottomrule
    \hline
\end{tabular}
}
\label{tab:Dataset}
\end{table}

Four evaluation metrics are used, including S-measure (S)\cite{fan2017structure}, max F-measure ($F_{\beta}$)\cite{achanta2009frequency}, max E-measure ($E_{\xi}$)\cite{fan2018enhanced},  and mean absolute error (M)\cite{perazzi2012saliency}.
\subsection{Implementation details}
The code is implemented in Pytorch toolbox on a PC with an NVIDIA RTX 3090 GPU.
SAM uses ViT-L version, so $L$ is 24.
The modal-aware modulation is only conducted in the last half layers.
During training, each image is uniformly resized to 1024$\times$1024. No data augmentation strategy is adopted. The batch size is 1 and the initial learning rate is 5e-5.  The model converges around 20 epochs by Adam optimizer \cite{kingma2014adam}. The training time is about 10 hours.

\subsection{Comparison with state-of-the-art methods}
\subsubsection{Quantitative comparisons}
The comparison results in RGB, RGB-D, RGB-T, and V-D-T SSSOD are shown in Table \ref{tab:RGBCom}, Table \ref{tab:RGBDCom}, Table \ref{tab:RGBTCom}, and  Table \ref{tab:VDTCom}, respectively.
In each table, we list some fully supervised methods, unsupervised methods, point supervised methods, box supervised methods, category supervised methods and some scribble supervised methods.
Meanwhile, we also collect the overall performances on multiple testing datasets in the ``Overall" column in terms of 5 datasets in RGB, 7 datasets in RGB-D, and 3 datasets in RGB-T.
For a fair comparison, the saliency maps of other methods are provided by authors or generated by the released codes and we evaluate them with the same evaluation code.

\begin{figure}[htp!]
	\centering
\begin{tabular}{c}
\includegraphics[width=1\linewidth]{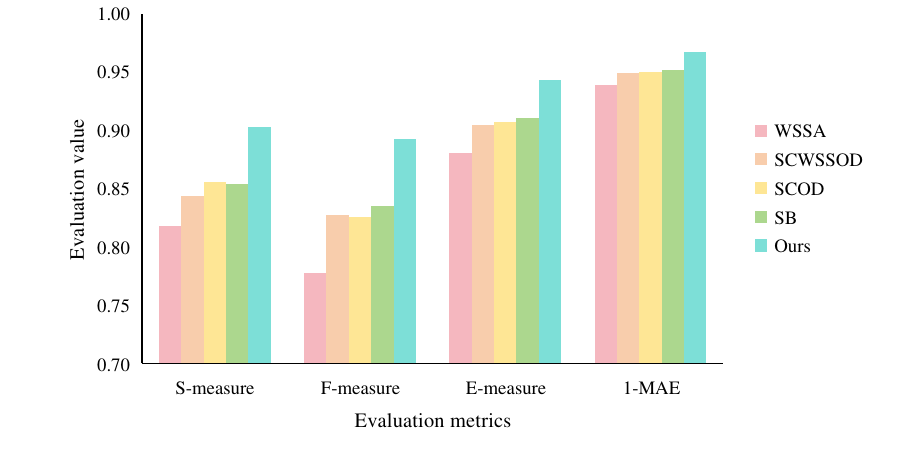}\\
\small(a) RGB scribble supervised SODs\\
\includegraphics[width=1\linewidth]{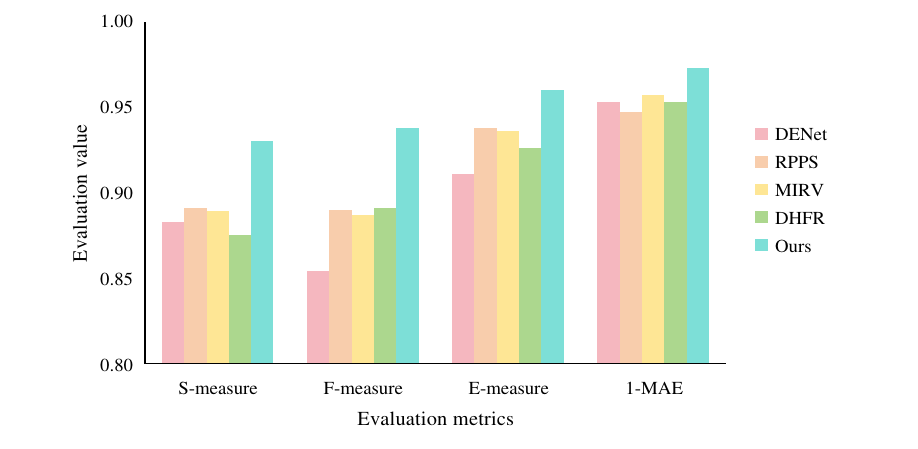}\\
\small(b) RGB-D scribble supervised SODs\\
\includegraphics[width=1\linewidth]{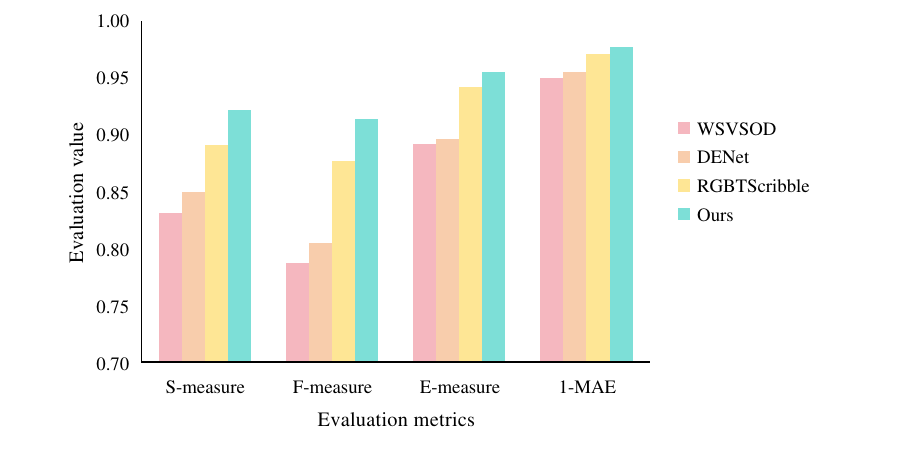}\\
\small(c) RGB-T scribble supervised SODs\\
\end{tabular}
\caption{Comparisons of overall performance in RGB, RGB-D, and RGB-T scribble supervised salient object detection methods. No V-D-T SSSOD methods are compared because we are the first to introduce the scribble supervised method in V-D-T SOD.  \label{fig:OverallCompare}}
\end{figure}

From the experimental results shown in all the datasets and the overall performance, our model consistently achieves non-trivial performance among scribble supervised methods.
In terms of the overall performance, our model improves the MAE evaluation metric about 0.02 in RGB and RGB-D dataset, and 0.01 in RGB-T dataset. As we are the first to achieve V-D-T scribble supervised method, no comparison method can be obtained. For the clearer demonstration, we present the overall performance comparisons in Fig \ref{fig:OverallCompare}.  From the figure, we can find our model significantly outperforms competitors.

Furthermore, our method is also very close to fully supervised methods and only lower than the best fully supervised method in MAE, e.g., RGB BBRF \cite{ma2023boosting}, RGBT PRLNet \cite{zhou2023position}, V-D-T QSFNet \cite{bao2024quality} about 0.002,  0.001, and 0.001, respectively. It is worth mentioning that all metric values on overall performance of  RGB-D SSSOD have surpassed those of fully supervised methods. The remarkable improvement benefits from the strong segmentation ability of fine-tuned SAM by  modal-aware modulators and the knowledge transfer from prompt decoder to no prompt decoder.

Last, our method is compared with some other weakly supervised methods, such as box \cite{liu2021weakly}, point \cite{gao2022weakly}, category \cite{wang2017learning,li2018weakly,zeng2019multi,piao2021mfnet,piao2022noise,li2021joint},
and some unsupervised unified methods \cite{yuan2024unified}. Existing methods constructed on the convolution neural networks can't win over our method based on the large model SAM with transformer backbones and our proposed modulation including modality-aware modulators and the siamese decoder.

\begin{table*}[!htp]
\caption{Quantitative comparisons on five RGB datasets and their overall performance. In fully supervised methods, the best results are underlined. In scribble supervised methods, the best results are bold. Pink background indicates our overall performance. `-' indicates no evaluation metrics report. `Category*' denotes category+caption supervision.}
\centering
\resizebox{1\linewidth}{!}
{
   \begin{tabular}{c|c|c|cccc|cccc|cccc}
\hline\toprule
   \multirow{2}{*}{\centering Methods} &\multirow{2}{*}{\centering Source} & \multirow{2}{*}{\centering Supervision} &\multicolumn{4}{c|}{\centering \textbf{Overall}}&
   \multicolumn{4}{c|}{\centering ECSSD} & \multicolumn{4}{c}{\centering DUTS-Test}  \\
     &&& S$\uparrow$
     & F$_\beta$ $\uparrow$ &$E_{\xi}\uparrow$
     & M$\downarrow$ & S$\uparrow$
     & F$_\beta$ $\uparrow$ &$E_{\xi}\uparrow$
     & M$\downarrow$ & S$\uparrow$
     & F$_\beta$$\uparrow$&$E_{\xi}\uparrow$ & M$\downarrow$
     \\
    \midrule

BIPG \cite{yao2021boundary}&TMM21&Full
&[.888]&[.874]&[.927]&[.038]
&.929&.945&.960&.029
&.896&.886&.937&.033\\
DPNet \cite{wu2022salient}&TIP22&Full
&[.867]&[.837]&[.910]&[.046]
&.918&.926&.951&.037
&.869&.841&.914&.043
\\
EDN \cite{wu2022edn}&TIP22&Full
&[\underline{.902}]&[.887]&[.938]&[.034]
&.938&.948&.965&.027
&\underline{.909}&.898&.948&.030\\
BBRF \cite{ma2023boosting}&TIP23 &Full
&[.900]&[\underline{.892}]&[\underline{.941}]&[\underline{.031}]
&\underline{.939}&\underline{.957}&\underline{.972}&\underline{.022}
&\underline{.909}&\underline{.906}&\underline{.952}&\underline{.025}\\
MENet \cite{wang2023pixels}&CVPR23&Full
&[.893]&[.876]&[.929]&[.034]
&.928&.939&.956&.031
&.905&.895&.944&.028\\
AiOSOD \cite{jia2023all}&Arxiv24&Full
&[.882]&[.859]&[.923]&[.044]
&.928&.939&.962&.034
&.882&.856&.927&.041\\
								
\midrule
A2Sv3 \cite{yuan2024unified}&IJCAI24&Un
&[.856]&[.839]&[.911]&[.048]
&.905&.926&.952&.038
&.848&.827&.908&.047\\
PSOD \cite{gao2022weakly}&AAAI22&Point
&[.861]&[.825]&[.895]&[.048]			
&.914&.921&.924&.036
&.853&.811&.887&.045\\
SBBs \cite{liu2021weakly}&TIP21&Box
&[.807]&[.756]&[.868]&[.068]			
&.851&.855&.894&.072
&.789&.722&.851&.073
\\
WSS \cite{wang2017learning}&CVPR17&Category
&[.764]&[.696]&[.818]&[.099]
&.811&.823&.869&.104
&.748&.654&.795&.100
\\
ASMO \cite{li2018weakly}&AAAI18&Category
&[.731]&[.638]&[.780]&[.111]			
&.802&.797&.853&.110
&.697&.614&.772&.116
\\
MSW \cite{zeng2019multi}&CVPR19&Category*
&[.779]&[.707]&[.823]&[.097]		
&.827&.840&.884&.096
&.759&.684&.814&.091
\\
MFNet \cite{piao2021mfnet}&ICCV21&Category
&[.787]&[.739]&[.851]&[.077]			
&.834&.854&.885&.084
&.775&.710&.839&.076
\\
NSAL \cite{piao2022noise}&TMM22&Category
&[.792]&[.749]&[.855]&[.074]			
&.834&.856&.884&.077
&.781&.730&.849&.073
\\
\midrule
WSSA \cite{zhang2020weakly}&CVPR20&Scribble
&[.818]&[.778]&[.881]&[.061]
&.865&.871&.920&.059
&.804&.755&.873&.062
\\
SCWSSOD \cite{yu2021structure}&AAAI21&Scribble
&[.844]&[.827]&[.905]&[.051]
&.882&.902&.932&.049
&.841&.823&.907&.049
\\
SCOD \cite{he2023}&AAAI23&Scribble
&[.856]&[.826]&[.907]&[.050]
&.902&.915&.946&.040
&.854&.822&.910&.047
\\
SB \cite{xu2023synthesize}&TMM23&Scribble
&[.854]&[.835]&[.902]&[.048]
&.890&.906&.917&.046
&.853&.834&.901&.045
\\
Ours&TMM24&Scribble
&\colorbox[RGB]{255,230,230}{[\textbf{.903}]}&\colorbox[RGB]{255,230,230}{[\textbf{.893}]}&\colorbox[RGB]{255,230,230}{[\textbf{.943}]}&\colorbox[RGB]{255,230,230}{[\textbf{.033}]}
&\textbf{.942}&\textbf{.954}&\textbf{.972}&\textbf{.023}
& \textbf{.904} &\textbf{.898}&\textbf{.946}&\textbf{.030}\\

   \bottomrule
   \toprule
   \multirow{2}{*}{\centering Methods} &\multirow{2}{*}{\centering Source} & \multirow{2}{*}{\centering Supervision} & \multicolumn{4}{c|}{\centering HKU-IS}&\multicolumn{4}{c|}{\centering DUT-OMRON} & \multicolumn{4}{c}{\centering PASCAL-S} \\

     &&& S$\uparrow$
     & F$_\beta$ $\uparrow$ &$E_{\xi}\uparrow$
     & M$\downarrow$ & S$\uparrow$
     & F$_\beta$ $\uparrow$ &$E_{\xi}\uparrow$
     & M$\downarrow$ & S$\uparrow$
     & F$_\beta$$\uparrow$&$E_{\xi}\uparrow$ & M$\downarrow$\\
    \midrule
BIPG \cite{yao2021boundary}&TMM21&Full
&.924&.936&.963&.025
&.845&.797&.882&.051
&.861&.859&.910&.060\\
DPNet \cite{wu2022salient}&TIP22&Full
&.905&.911&.950&.034
&.825&.754&.868&.056
&.848&.833&.891&.071\\
EDN \cite{wu2022edn}&TIP22&Full
&\underline{.934}&.942&.969&.023
&\underline{.865}&.820&\underline{.899}&.045
&.877&.874&.916&.056\\
BBRF \cite{ma2023boosting}&TIP23&Full
&.932&\underline{.946}&\underline{.972}&\underline{.020}
&.861&\underline{.822}&\underline{.899}&\underline{.044}
&\underline{.878}&\underline{.879}&\underline{.927}&\underline{.049}\\
MENet \cite{wang2023pixels}&CVPR23&Full
&.927&.939&.965&.023
&.850&.792&.879&.045
&.872&.869&.915&.054\\
AiOSOD\cite{jia2023all}&Arxiv24&Full
&.918&.927&.961&.033
&.845&.789&.882&.055
&.863&.853&.910&.064\\
\midrule
A2Sv3\cite{yuan2024unified}&IJCAI24&Un
   &.898&.913&.954&.033
   &.821&.771&.872&.062
   &.843&.843&.900&.067
   \\

PSOD \cite{gao2022weakly}&AAAI22&Point
&.902&.907&.952&.032
&.824&.748&.854&.064
&.853&.831&.857&.065\\
SBBs \cite{liu2021weakly}&TIP21&Box
&.854&.843&.920&.056
&.776&.695&.835&.074
&-&-&-&-\\
WSS \cite{wang2017learning}&CVPR17&Category
&.822&.821&.896&.079
&.725&.603&.768&.109
&.744&.715&.791&.139\\
ASMO \cite{li2018weakly}&AAAI18&Category
&-&-&-&-
&.752&.622&.776&.101
&.717&.693&.772&.149\\
MSW \cite{zeng2019multi}&CVPR19&Category*
&.818&.814&.895&.084
&.756&.609&.763&.109
&.768&.713&.790&.133\\
MFNet \cite{piao2021mfnet}&ICCV21&Category
&.846&.851&.921&.059
&.742&.646&.803&.087
&.770&.751&.817&.115\\
NSAL \cite{piao2022noise}&TMM22&Category
&.854&.864&.923&.051
&.745&.648&.801&.088
&.768&.756&.822&.110\\
\midrule

WSSA \cite{zhang2020weakly}&CVPR20&Scribble
&.865&.865&.931&.047
&.785&.708&.841&.068
&.797&.780&.861&.092\\
SCWSSOD \cite{yu2021structure}&AAAI21&Scribble
&.882&.898&.943&.038
&.812&.756&.868&.060
&.820&.821&.881&.078\\
SCOD\cite{he2023}&AAAI23&Scribble
&.904&.909&.954&.032
&.809&.739&.856&.068
&.841&.835&.898&.069
\\
SB \cite{xu2023synthesize}&TMM23&Scribble
&.895&.906&.949&.034
&.818&.762&.868&.060
&.825&.827&.853&.073\\
Ours&TMM24&Scribble
&\textbf{.934}&\textbf{.942}&\textbf{.970}&\textbf{.022}
&\textbf{.873}&\textbf{.838}&\textbf{.915}&\textbf{.044}
&\textbf{.876}&\textbf{.875}&\textbf{.921}&\textbf{.055}\\

     \bottomrule
    \hline
\end{tabular}
}
\label{tab:RGBCom}
\end{table*}

\begin{table*}[!htp]
\caption{Quantitative comparisons on seven RGB-D datasets and their overall performance. In fully supervised methods, the best results are underlined. In scribble supervised methods, the best results are bold. Pink background indicates our overall performance. `-' indicates no evaluation metrics report. `Category*' denotes category+caption supervision.}
\centering
\resizebox{1\linewidth}{!}
{
   \begin{tabular}{c|c|c|cccc|cccc|cccc|cccc}
\hline\toprule
   \multirow{2}{*}{\centering Methods}&\multirow{2}{*}{\centering Source} & \multirow{2}{*}{\centering Supervision} &\multicolumn{4}{c|}{\centering \textbf{Overall}}&
   \multicolumn{4}{c|}{\centering NLPR} & \multicolumn{4}{c|}{\centering NJU2K} & \multicolumn{4}{c}{\centering STERE}  \\
     &&& S$\uparrow$
     & F$_\beta$ $\uparrow$ &$E_{\xi}\uparrow$
     & M$\downarrow$ & S$\uparrow$
     & F$_\beta$ $\uparrow$ &$E_{\xi}\uparrow$
     & M$\downarrow$ & S$\uparrow$
     & F$_\beta$$\uparrow$&$E_{\xi}\uparrow$ & M$\downarrow$
     & S$\uparrow$
     & F$_\beta$$\uparrow$ &$E_{\xi}\uparrow$& M$\downarrow$ \\
    \midrule
    JL-DCF \cite{fu2021siamese}&TPAMI21&Full
    &[.905]&[.904]&[.949]&[.040]
    &.926&.917&.964&.023
    &.911&.913&.948&.040
    &.911&.907&.949&.039\\
    SwinNet \cite{liu2021swinnet}&TCSVT22&Full
    &[.921]&[.925]&[.956]&[.031]
    &.941&\underline{.936}&\underline{.974}&.018
    &.935&.938&.963&.027
    &.919&.918&.956&.033
    \\
    PCIR-Net \cite{cong2023point}&ACM MM23&Full
    &[.915]&[.920]&[.952]&[.033]
    &.935&.931&.970&.019
    &.927&.931&.958&.029
    &.920&.920&.957&.031
    \\
    HRTransNet \cite{tang2022hrtransnet}&TCSVT23&Full
    &[.920]&[.926]&[.956]&[.030]
    &\underline{.942}&\underline{.936}&\underline{.974}&\underline{.016}
    &.933&.939&.963&.026
    &.921&.919&.956&\underline{.030}
    \\
    HiDANet \cite{wu2023hidanet}&TIP23&Full
    &[.910]&[.924]&[.944]&[.034]
    &.930&.929&.961&.021
    &.926&.939&.954&.029
    &.911&.921&.946&.035
    \\
    CATNet \cite{sun2023catnet}&TMM23&Full
    &[.922]&[.927]&[.958]&[.029]
    &.940&.934&.972&.018
    &.932&.937&.960&.026
    &.921&\underline{.922}&\underline{.958}&\underline{.030}
    \\
AiOSOD \cite{jia2023all}&Arxiv24&Full
&[.915]&[.915]&[.951]&[.036]
&.927&.911&.958&.025
&.925&.923&.957&.033
&.915&.909&.950&.038\\
    DFormer \cite{yin2023dformer}&ICLR24&Full
    &[\underline{.925}]&[\underline{.929}]&[\underline{.959}]&[\underline{.027}]
    &\underline{.942}&\underline{.936}&.973&\underline{.016}
    &\underline{.937}&\underline{.943}&\underline{.967}&\underline{.023}
    &\underline{.923}&.920&.956&\underline{.030}
    \\  	

\midrule

    A2Sv3 \cite{yuan2024unified}&IJCAI24&Un
    &[.887]&[.889]&[.934]&[.042]
    &.905&.892&.949&.028
    &.881&.882&.924&.049
    &.892&.895&.941&.040\\
    JSM \cite{li2021joint}&NeurIPS21&Category*
    &[.745] &[.721] &[.820] &[.115]
    &.805&.771&.875&.060
    &.713&.711&.801&.133						
    &.782&.772&.852&.096\\
   \midrule
    DENet \cite{xu2022weakly}&TIP22&Scribble
    &[.883]&[.854]&[.911]&[.047]
    &.913&.874&.936&.028
    &.899&.878&.922&.044
    &.887&.851&.914&.044
    \\
    DHFR \cite{liu2023deep}&TIP23&Scribble
    &[.875]&[.891]&[.926]&[.047]
    &.904&.901&.950&.027
    &.893&.901&.936&.040
    &.884&.896&.935&.043
    \\
    RPPS\cite{li2023robust}&TPAMI24&Scribble
    &[.891]&[.890]&[.938]&[.053]
    &.911&.896&.953&.033
    &.893&.899&.940&.054
    &.882&.883&.931&.061
    \\
    MIRV \cite{li2023mutual}&TCSVT24&Scribble
    &[.889]&[.887]&[.936]&[.043]
    &.914&.904&.957&.025
    &.890&.890&.934&.046
    &.891&.888&.939&.042\\

    Ours&TMM24&Scribble
    &\colorbox[RGB]{255,230,230}{[\textbf{.930}]}&\colorbox[RGB]{255,230,230}{[\textbf{.938}]}&\colorbox[RGB]{255,230,230}{[\textbf{.960}]}&\colorbox[RGB]{255,230,230}{[\textbf{.027}]}
    &\textbf{.936}&\textbf{.932}&\textbf{.965}&\textbf{.018}
    &\textbf{.932}&\textbf{.941}&\textbf{.957}&\textbf{.028}
    &\textbf{.935}&\textbf{.947}&\textbf{.966}&\textbf{.025}
    \\
   \bottomrule
   \toprule
   \multirow{2}{*}{\centering Methods} &\multirow{2}{*}{\centering Source}& \multirow{2}{*}{\centering Supervision} & \multicolumn{4}{c|}{\centering SIP}&\multicolumn{4}{c|}{\centering DES} & \multicolumn{4}{c|}{\centering LFSD} & \multicolumn{4}{c}{\centering SSD}\\
     &&& S$\uparrow$
     & F$_\beta$ $\uparrow$ &$E_{\xi}\uparrow$
     & M$\downarrow$ & S$\uparrow$
     & F$_\beta$ $\uparrow$ &$E_{\xi}\uparrow$
     & M$\downarrow$ & S$\uparrow$
     & F$_\beta$$\uparrow$&$E_{\xi}\uparrow$ & M$\downarrow$& S$\uparrow$
     & F$_\beta$$\uparrow$&$E_{\xi}\uparrow$ & M$\downarrow$\\
    \midrule
    JL-DCF \cite{fu2021siamese}&TPAMI21&Full
    &.892&.900&.949&.046
    &.936&.929&.975&.021
    &.863&.862&.900&.071
    &.860&.833&.902&.053\\
    SwinNet\cite{liu2021swinnet}&TCSVT22&Full
    &.911&.927&.950&.035
    &.945&.945&.979&.016
    &.886&.889&.921&.059
    &.892&.878&.925&.040
    \\
     PCIR-Net \cite{cong2023point}&ACM MM23&Full
    &.899&.915&.939&.040
    &.942&.942&.977&.016
    &.888&\underline{.894}&.923&.052
    &.874&.862&.923&.047
    \\
    HRTransNet\cite{tang2022hrtransnet}&TCSVT23&Full
    &.909&.929&.949&.035
    &.947&\underline{.952}&.982&.014
    &.877&.872&.912&.058
    &\underline{.898}&\underline{.890}&\underline{.934}&\underline{.035}
    \\
    HiDANet \cite{wu2023hidanet}&TIP23&Full
    &.892&.919&.927&.043
    &.946&\underline{.952}&.980&\underline{.013}
    &-&-&-&-
    &.871&.854&.915&.047
    \\
    CATNet \cite{sun2023catnet}&TMM23&Full
    &.911&.928&.952&.034
    &.942&.940&.977&.017
    &\underline{.894}&.892&\underline{.928}&\underline{.051}
    &-&-&-&-\\
    AiOSOD \cite{jia2023all}&Arxiv24&Full
    &.907&.922&.949&.038
    &.946&.940&.980&.018
    &.886&.886&.916&.061
    &.885&.866&.931&.044\\
    DFormer \cite{yin2023dformer}&ICLR24&Full
    &\underline{.915}&\underline{.930}&\underline{.953}&\underline{.032}
    &\underline{.948}&\underline{.952}&\underline{.984}&\underline{.013}	
    &-&-&-&-
    &.894&.885&\underline{.934}&\underline{.035}\\			

\midrule
    A2Sv3 \cite{yuan2024unified} &IJCAI24&Un
    &.881&.889&.932&.044
    &.904&.902&.939&.025
    &.863&.874&.909&.065
    &.859&.845&.916&.050\\
    JSM \cite{li2021joint}&NeurIPS21&Category*
    &.697&.645&.774&.146
    &.820&.807&.875&.058
    &.767&.781&.833&.127						
    &.652&.617&.751&.157\\
\midrule

    DENet\cite{xu2022weakly}&TIP22&Scribble
    &.865&.841&.900&.056
    &.916&.887&.939&.029
    &.832&.814&.860&.085
    &.850&.799&.873&.061
    \\
    DHFR \cite{liu2023deep}&TIP23&Scribble
    &.843&.874&.902&.064
    &.915&.930&.956&.022
    &.855&.878&.904&.066
    &.858&.869&.911&.051
    \\
    RPPS \cite{li2023robust}&TPAMI24&Scribble
    &.892&.892&.939&.050
    &.919&.914&.962&.027
    &.868&.877&.913&.078
    &.868&.856&.919&.056
    \\
    MIRV\cite{li2023mutual}&TCSVT24&Scribble
    &.876&.879&.928&.049
    &.928&.921&.965&.019
    &.849&.844&.889&.072
    &.891&.873&.934&.042\\
%
    Ours&TMM24&Scribble
    &\textbf{.927}&\textbf{.938}&\textbf{.958}&\textbf{.028}
    &\textbf{.930}&\textbf{.933}&\textbf{.967}&\textbf{.017}
    &\textbf{.891}&\textbf{.892}&\textbf{.920}&\textbf{.057}
    &\textbf{.902}&\textbf{.900}&\textbf{.945}&\textbf{.034}
    \\

     \bottomrule
    \hline
\end{tabular}
}
\label{tab:RGBDCom}
\end{table*}

\begin{table*}[!htp]
\caption{Quantitative comparisons on three RGB-T datasets and their overall performance. In fully supervised methods, the best results are underlined. In scribble supervised methods, the best results are bold. Pink background indicates our overall performance.}
\centering
\resizebox{1\linewidth}{!}
{
   \begin{tabular}{c|c|c|cccc|cccc|cccc|cccc}
\hline\toprule
   \multirow{2}{*}{\centering Methods}&\multirow{2}{*}{\centering Source} & \multirow{2}{*}{\centering Supervision} &\multicolumn{4}{c|}{\centering \textbf{Overall}}&\multicolumn{4}{c|}{\centering VT821 } & \multicolumn{4}{c|}{\centering VT1000 } & \multicolumn{4}{c}{\centering VT5000 }\\
     &&& S$\uparrow$
     & F$_\beta$ $\uparrow$ &$E_{\xi}\uparrow$
     & M$\downarrow$& S$\uparrow$
     & F$_\beta$ $\uparrow$ &$E_{\xi}\uparrow$
     & M$\downarrow$ & S$\uparrow$
     & F$_\beta$ $\uparrow$ &$E_{\xi}\uparrow$
     & M$\downarrow$ & S$\uparrow$
     & F$_\beta$$\uparrow$&$E_{\xi}\uparrow$ & M$\downarrow$\\
    \midrule

    TNet\cite{cong2022does}&TMM22&Full
    &[.904]&[.894]&[.944]&[.030]
    &.899&.888&.938&.030
    &.929&.930&.966&.021
    &.895&.881&.937&.033\\
    HRTransNet \cite{tang2022hrtransnet}&TCSVT23&Full
    &[.917]&[.909]&[.958]&[.023]
    &.906&.888&.941&.026
    &.938&.941&.975&.017
    &.912&.903&.956&.025\\
    CAVER \cite{pang2023caver}&TIP23&Full
    &[.908]&[.894]&[.949]&[.025]
    &.898&.877&.934&.027
    &.938&.939&.973&.017
    &.899&.882&.944&.028\\
    WaveNet\cite{zhou2023wavenet}&TIP23&Full
    &[.919]&[.907]&[.955]&[.023]
    &.912&.895&.943&\underline{.024}
    &\underline{.945}&.945&.977&.015
    &.911&.896&.950&.026\\
    PRLNet \cite{zhou2023position}&TIP23&Full
    &[\underline{.926}]&[\underline{.923}]&[\underline{.965}]&[.022]
    &\underline{.917}&\underline{.910}&\underline{.954}&.025
    &.944&\underline{.950}&\underline{.980}&.016
    &\underline{.921}&\underline{.916}&\underline{.962}&\underline{.023}\\
AiOSOD \cite{jia2023all}&Arxiv24&Full
    &[.908]&[.892]&[.946]&[.030]
    &.905&.882&.937&.028
    &.941&.941&.974&.020
    &.895&.875&.938&.035\\
UniTR \cite{guo2024unitr}&TMM24&Full
    &[.913]&[.914]&[.958]&[\underline{.021}]
    &.901&.898&.941&.025
    &.938&.939&.975&\underline{.014}
    &.907&.910&.956&\underline{.023}\\
\midrule
A2Sv3 \cite{yuan2024unified}&IJCAI24&Un
    &[.886]&[.870]&[.932]&[.035]
    &.882&.866&.922&.041
    &.923&.921&.959&.023
    &.872&.851&.924&.038\\
\midrule
    WSVSOD \cite{zhao2021weakly}&CVPR21&Scribble
    &[.831]&[.788]&[.892]&[.050]
    &.822&.762&.875&.052
    &.886&.870&.935&.035
    &.812&.763&.880&.055\\
    DENet \cite{xu2022weakly}&TIP22&Scribble
    &[.850]&[.805]&[.896]&[.045]
    &.813&.732&.839&.054
    &.906&.891&.947&.028
    &.839&.794&.895&.048\\
    RGBTScribble \cite{liu2023RGBTScribble}&ICME23&Scribble
    &[.892]&[.877]&[.942]&[.029]
    &.895&.878&.942&.027
    &.925&.922&.964&.020
    &.877&.859&.933&.033\\
    Ours&TMM24&Scribble
    &\colorbox[RGB]{255,230,230}{[\textbf{.922}]}&\colorbox[RGB]{255,230,230}{[\textbf{.914}]}&\colorbox[RGB]{255,230,230}{[\textbf{.955}]}&\colorbox[RGB]{255,230,230}{[\textbf{.023}]}
    &\textbf{.919}&\textbf{.896}&\textbf{.943}&\textbf{.023}
    &\textbf{.943}&\textbf{.942}&\textbf{.973}&\textbf{.015}
    &\textbf{.914}&\textbf{.909}&\textbf{.951}&\textbf{.026}\\
   \bottomrule
    \hline
\end{tabular}
}
\label{tab:RGBTCom}
\end{table*}

\begin{table}[!htp]
\caption{Quantitative comparisons on VDT-2048 dataset. In fully supervised methods, the best results are underlined. Our scribble supervised method is trained using our relabeled VDT-S SOD datasets.}
\centering
\resizebox{1\linewidth}{!}
{
   \begin{tabular}{ccccccc}
\hline\toprule
   \multirow{2}{*}{\centering Methods} &\multirow{2}{*}{\centering Source} & \multirow{2}{*}{\centering Supervision} &\multicolumn{4}{c}{\centering VDT-2048}\\

     &&& S$\uparrow$
     & F$_\beta$ $\uparrow$ &$E_{\xi}\uparrow$
     & M$\downarrow$\\
    \midrule

    HWSI  \cite{song2022novel}&TMECH23&Full
    &.932&.897&.981&.003\\
    TMNet \cite{wan2023tmnet}&PR24&Full
    &.933&.910&.989&.003\\
    FFANet \cite{zhou2024frequency}&PR24&Full
    &.935&.893&.989&.003\\
    MFFNet \cite{wan2023mffnet}&TMM23&Full
    &.936&.914&.990&\underline{.002}\\
    QSFNet \cite{bao2024quality}&TIP24&Full
    &\underline{.943}&\underline{.925}&\underline{.993}&\underline{.002}\\
\midrule
A2Sv3  \cite{yuan2024unified} &IJCAI24&Un
    &.716&.545&.734&.006\\
\midrule
    Ours&TMM24&Scribble
    &\colorbox[RGB]{255,230,230}{.916}&\colorbox[RGB]{255,230,230}{.884}&\colorbox[RGB]{255,230,230}{.978}&\colorbox[RGB]{255,230,230}{.003}
    \\
   \bottomrule
    \hline
\end{tabular}
}
\label{tab:VDTCom}
\end{table}

\subsubsection{Visual comparisons}
\begin{figure}[!htp]
  \centering
  \includegraphics[width=1\linewidth]{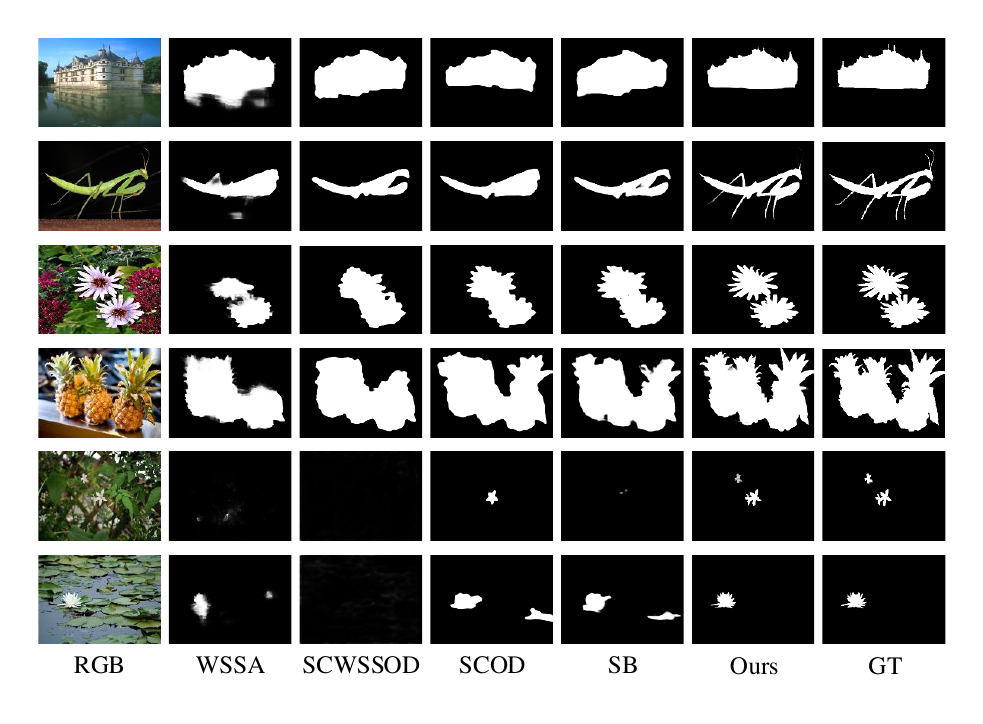}\\
  \caption{{Visual comparisons with RGB SSSOD competitors in some challenging cases: reflecting object (1st row), fine-grained object (2nd row), multiple objects (3rd-4th rows),  and small objects (5th-6th rows).}
  \label{fig:RGBVisual}}
\end{figure}

\begin{figure}[!htp]
  \centering
  \includegraphics[width=1\linewidth]{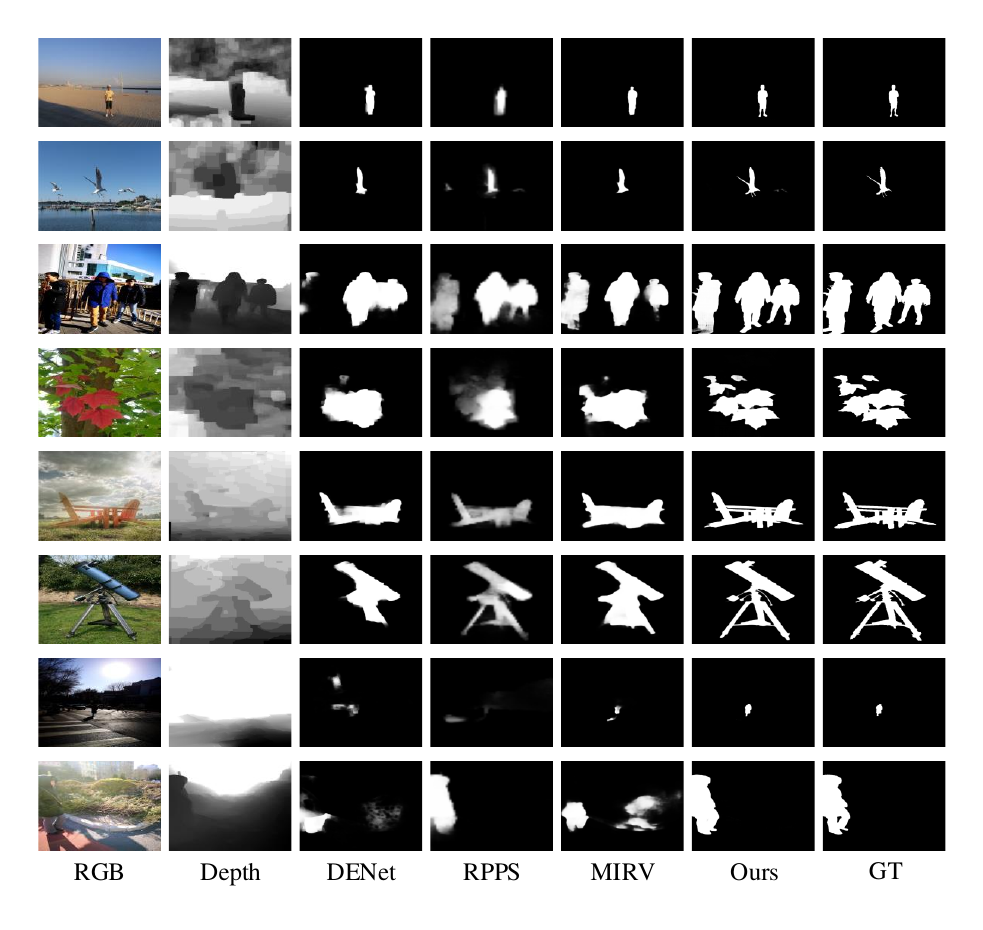}\\
  \caption{{Visual comparisons with RGB-D SSSOD competitors in some challenging cases: small objects (1st-2nd rows), multiple objects (3rd-4th rows), hollow objects (5th-6th rows), and light interference (7th-8th rows).}
  \label{fig:RGBDVisual}}
\end{figure}

\begin{figure}[!htp]
  \centering
  \includegraphics[width=1\linewidth]{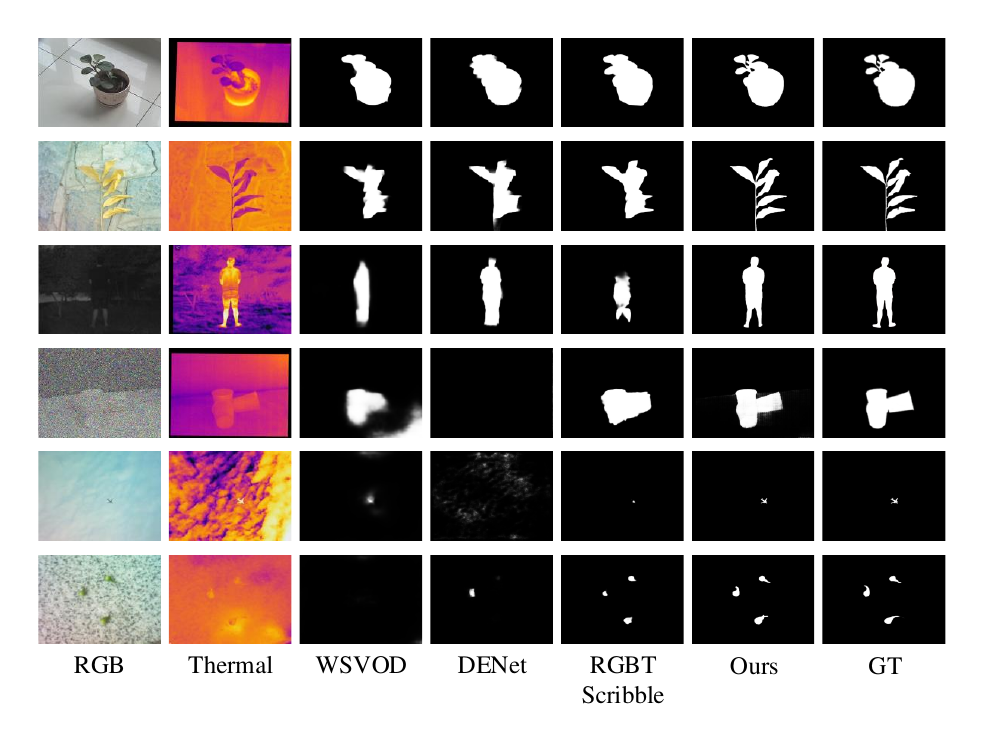}\\
  \caption{{Visual comparisons with RGB-T SSSOD competitors in some challenging cases:  fine-grained objects (1st-2nd rows), low-quality RGB images (3rd-4th rows), and small objects (5th-6th rows).}
  \label{fig:RGBTVisual}}
\end{figure}

\begin{figure}[!htp]
  \centering
  \includegraphics[width=1\linewidth]{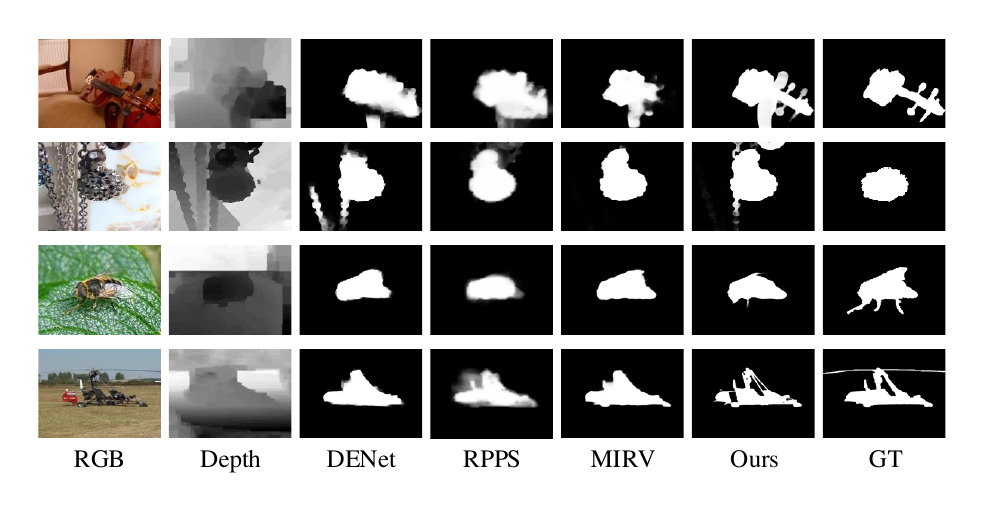}
  \caption{{Failure cases: severe background interference (1st-2nd rows), fine-grained details (3rd-4th rows).}
  \label{fig:Failure}}
\end{figure}

Fig \ref{fig:RGBVisual} shows some visual examples for  reflecting object, fine-grained object, multiple objects, and small objects in RGB SOD task.
Fig \ref{fig:RGBDVisual} shows some visual examples for small objects, multiple objects, hollow objects, and light interference in RGB-D SOD task.
Fig \ref{fig:RGBTVisual} shows some visual examples for fine-grained objects, low-quality RGB images, and small objects in RGB-T SOD task.
Our results are obviously better, which indicate the advantage of our model.
However, there are some failure cases. In the first two rows of Fig \ref{fig:Failure}, when foreground and background are similar and their depths are also close, some non-salient objects are detected. In the last two rows of Fig \ref{fig:Failure}, the fine-grained object parts, e.g., the legs of insects and the propeller, are missing. It can be inferred from the performances of the other scribble supervised methods that it is a great challenge to segment the complete object from sparse scribble labels.

\subsubsection{Complexity comparisons}
Take RGB-D SSSOD as an example, Table \ref{tab:CostComp} presents complexity comparison. From the result, it can be observed that the computation cost of our model is larger,  because the core backbone SAM is a large model. However, we also find the trainable parameters are small because modal-aware modulators use the LoRAs  with low-rank properties and the siamese decoder is two lightweight SAM decoders.
\begin{table}[!htp]
\caption{Complexity comparisons with RGB-D SSSOD competitors in terms of input resolution, computation cost, and model size. `-' indicates no evaluation metrics report.}
\centering
\resizebox{1\linewidth}{!}
{
   \begin{tabular}{ccccc}
\toprule
    & DENet\cite{xu2022weakly}&RPPS \cite{li2023robust}&DHFR \cite{liu2023deep}&Ours \\

    \midrule
    Input Resolution&352$\times$352& 256$\times$256&224$\times$224&1024$\times$1024\\

    FLOPs (G) &95.27&23.10&-&2635.70\\

    Trainable Params (M) &37.14&53.39&54.5&9.24\\

    Overall Params (M) &37.14&53.39&54.5&303.24\\
   \bottomrule
    \hline
\end{tabular}
}
\label{tab:CostComp}
\end{table}

\subsection{Ablation studies}
All the ablation studies are taken on RGB-D SSSOD task.
\subsubsection{Ablation on different modulators}
As SAM is a large model with a large number of parameters, the cost of full-tuning model is high, which is not achieved by common single-card GPU. Among the prevalent fine-tuning techniques, adapter and LoRA are compared as the modulators for RGB-D SSSOD task.
From the results in ``Overall" column of Table \ref{tab:RGBDModulatorAblation}, we can see that fine-tuning is effective regardless adapter and LoRA, and LoRA is superior to adapter  in the performance.  Moreover, we list the number of trainable parameters when no modulators, using adapters, and using LoRAs. The number of   parameters in two decoders is 7.74M. It adds the number of   parameters in RGB and depth adapters or LoRAs and the sums are 13.81M and 9.24M, respectively. It manifests the effectiveness of LoRAs compared to adapters.
\begin{table*}[!htp]
\caption{Ablation study on RGB-D SOD datasets about the modulators. The best results are bold, and  pink background indicates our overall performance.}
\centering
\resizebox{1\linewidth}{!}
{
   \begin{tabular}{c|c|cccc|cccc|cccc|cccc}
\hline\toprule
   \multirow{2}{*}{\centering Methods} &Trainable &\multicolumn{4}{c|}{\centering \textbf{Overall}}&   \multicolumn{4}{c|}{\centering NLPR} & \multicolumn{4}{c|}{\centering NJU2K} & \multicolumn{4}{c}{\centering STERE}  \\
     &Params(M)
     & S$\uparrow$
     & F$_\beta$ $\uparrow$ &$E_{\xi}\uparrow$
     & M$\downarrow$ & S$\uparrow$
     & F$_\beta$ $\uparrow$ &$E_{\xi}\uparrow$
     & M$\downarrow$ & S$\uparrow$
     & F$_\beta$$\uparrow$&$E_{\xi}\uparrow$ & M$\downarrow$
     & S$\uparrow$
     & F$_\beta$$\uparrow$ &$E_{\xi}\uparrow$& M$\downarrow$ \\
    \midrule
    No modulator
    &7.74
    &[.912]&[.920]&[.949]&[.034]
    &.919&.920&.961&.023
    &.919&.921&.949&.034
    &.914&.921&.953&.034
    \\
    Adapter modulators
    &13.81
    &[.919]&[.923]&[.953]&[.031]
    &.930&.926&.963&.019
    &.929&.931&\textbf{.957}&\textbf{.028}
    &.928&.936&.962&.027
    \\
    LoRA modulators (Ours)
    &9.24
    &\colorbox[RGB]{255,230,230}{[\textbf{.930}]}&\colorbox[RGB]{255,230,230}{[\textbf{.938}]}&\colorbox[RGB]{255,230,230}{[\textbf{.960}]}&\colorbox[RGB]{255,230,230}{[\textbf{.027}]}
    &\textbf{.936}&\textbf{.932}&\textbf{.965}&\textbf{.018}
    &\textbf{.932}&\textbf{.941}&\textbf{.957}&\textbf{.028}
    &\textbf{.935}&\textbf{.947}&\textbf{.966}&\textbf{.025}
    \\
   \bottomrule
   \toprule
   \multirow{2}{*}{\centering Methods} &FLOPs&\multicolumn{4}{c|}{\centering SIP}&\multicolumn{4}{c|}{\centering DES} & \multicolumn{4}{c|}{\centering LFSD} & \multicolumn{4}{c}{\centering SSD}\\
   & (T) & S$\uparrow$
     & F$_\beta$ $\uparrow$ &$E_{\xi}\uparrow$
     & M$\downarrow$ & S$\uparrow$
     & F$_\beta$ $\uparrow$ &$E_{\xi}\uparrow$
     & M$\downarrow$ & S$\uparrow$
     & F$_\beta$$\uparrow$&$E_{\xi}\uparrow$ & M$\downarrow$& S$\uparrow$
     & F$_\beta$$\uparrow$&$E_{\xi}\uparrow$ & M$\downarrow$\\
    \midrule
    No modulator
    &2.63
    &.904&.921&.942&.038
    &\textbf{.945}&\textbf{.944}&\textbf{.981}&\textbf{.013}
    &.879&.883&.913&.063
    &.884&.884&.938&.041\\
    Adapter modulators

    &2.65
    &.905&.911&.943&.038
    &.929&.927&.969&.018
    &.884&.886&.912&\textbf{.057}
    &.886&.884&.931&.039\\
    LoRA modulators (Ours)
    &2.64
    &\textbf{.927}&\textbf{.938}&\textbf{.958}&\textbf{.028}
    &.930&.933&.967&.017
    &\textbf{.891}&\textbf{.892}&\textbf{.920}&\textbf{.057}
    &\textbf{.902}&\textbf{.900}&\textbf{.945}&\textbf{.034}\\
     \bottomrule
    \hline
\end{tabular}
}
\label{tab:RGBDModulatorAblation}
\end{table*}

\subsubsection{Ablation on decoder}
To verify the effectiveness of our siamese decoder, we conduct the ablation on the decoder. Since SAM provide the prompt interface, scribble can be regarded as both supervision signals and prompts. However, there is no prompt in the testing stage. Therefore,  the model which is trained with scribble prompt and tested with no prompt in the first row of Table \ref{tab:RGBDDecoderAblation} are obviously inferior to the model which is trained and tested both with no prompt in the second row. It reveals that there is a gap between training with prompts and testing with no prompt. Furthermore, the majority  of the results in the second row of Table \ref{tab:RGBDDecoderAblation} are  worse than ours in the third row, which indicating that our siamese decoder can transfer the knowledge of the trained model with prompts to the testing model with no prompt and deliver the better performance in the testing stage.
\begin{table*}[!htp]
\caption{Ablation study on RGB-D SOD datasets about the decoder. The best results are bold, and  pink background indicates our overall performance.}
\centering
\resizebox{1\linewidth}{!}
{
   \begin{tabular}{c|cccc|cccc|cccc|cccc}
\hline\toprule
   \multirow{2}{*}{\centering Methods} &\multicolumn{4}{c|}{\centering \textbf{Overall}}&   \multicolumn{4}{c|}{\centering NLPR} & \multicolumn{4}{c|}{\centering NJU2K} & \multicolumn{4}{c}{\centering STERE}  \\
     & S$\uparrow$
     & F$_\beta$ $\uparrow$ &$E_{\xi}\uparrow$
     & M$\downarrow$ & S$\uparrow$
     & F$_\beta$ $\uparrow$ &$E_{\xi}\uparrow$
     & M$\downarrow$ & S$\uparrow$
     & F$_\beta$$\uparrow$&$E_{\xi}\uparrow$ & M$\downarrow$
     & S$\uparrow$
     & F$_\beta$$\uparrow$ &$E_{\xi}\uparrow$& M$\downarrow$ \\
    \midrule
    scribble prompt train\& no prompt test
    &[.915]&[.916]&[.949]&[.035]
    &.928&.916&.958&.022
    &.923&.927&.954&.033
    &.929&.931&.959&.027
    \\
    no prompt train\& no prompt test
    &[.920]&[.928]&[.955]&[.031]
    &.932&.925&.964&.020
    &.927&.937&\textbf{.957}&.031
    &.925&.942&\textbf{.966}&.029
    \\
    siamese decoder (Ours)
    &\colorbox[RGB]{255,230,230}{[\textbf{.930}]}&\colorbox[RGB]{255,230,230}{[\textbf{.938}]}&\colorbox[RGB]{255,230,230}{[\textbf{.960}]}&\colorbox[RGB]{255,230,230}{[\textbf{.027}]}
    &\textbf{.936}&\textbf{.932}&\textbf{.965}&\textbf{.018}
    &\textbf{.932}&\textbf{.941}&\textbf{.957}&\textbf{.028}
    &\textbf{.935}&\textbf{.947}&\textbf{.966}&\textbf{.025}
    \\
   \bottomrule
   \toprule
   \multirow{2}{*}{\centering Methods} & \multicolumn{4}{c|}{\centering SIP}&\multicolumn{4}{c|}{\centering DES} & \multicolumn{4}{c|}{\centering LFSD} & \multicolumn{4}{c}{\centering SSD}\\
     & S$\uparrow$
     & F$_\beta$ $\uparrow$ &$E_{\xi}\uparrow$
     & M$\downarrow$ & S$\uparrow$
     & F$_\beta$ $\uparrow$ &$E_{\xi}\uparrow$
     & M$\downarrow$ & S$\uparrow$
     & F$_\beta$$\uparrow$&$E_{\xi}\uparrow$ & M$\downarrow$& S$\uparrow$
     & F$_\beta$$\uparrow$&$E_{\xi}\uparrow$ & M$\downarrow$\\
    \midrule
    scribble prompt train\& no prompt test
    &.896&.901&.934&.047
    &.921&.917&.965&.022
    &.877&.877&.908&.064
    &.892&.880&.943&.039\\
    no prompt train\& no prompt test
    &.911&.916&.944&.036
    &\textbf{.931}&.929&\textbf{.968}&.018
    &.885&.887&.917&.060
    &.892&.891&.934&.038\\
    siamese decoder (Ours)
    &\textbf{.927}&\textbf{.938}&\textbf{.958}&\textbf{.028}
    &.930&\textbf{.933}&.967&\textbf{.017}
    &\textbf{.891}&\textbf{.892}&\textbf{.920}&\textbf{.057}
    &\textbf{.902}&\textbf{.900}&\textbf{.945}&\textbf{.034}\\
     \bottomrule
\end{tabular}
}
\label{tab:RGBDDecoderAblation}
\end{table*}

\subsubsection{Ablation on different modalities}
RGB-D SOD task is inspired by the fact that depth can provide the spatial cue when RGB image is not enough to provide effective information in some challenging scenes, for example, foreground and background are similar, and scenes are cluttered.
The existing methods combine RGB and depth features to remedy the defect of unimodal input.
To show the usefulness of depth cue in our model, we compare the performance with RGB modality and that with RGB and depth modalities.
From the results in Table \ref{tab:RGBDModalAblation}, we find depth modality plays an auxiliary role because the performance in MAE evaluation metric is slightly improved about 0.004 in overall. Moreover, RGB's performance has surpassed a few of existing fully supervised RGB-D methods in Table \ref{tab:RGBDCom}.
The above results consistently give the hint that the features extracted from SAM is very good so that the role of depth information is minor. 
\begin{table*}[!htp]
\caption{Ablation study on RGB-D SOD datasets about the modalities. The best results are bold, and  pink background indicates our overall performance.}
\centering
\resizebox{1\linewidth}{!}
{
   \begin{tabular}{c|cccc|cccc|cccc|cccc}
\hline\toprule
   \multirow{2}{*}{\centering Methods} &\multicolumn{4}{c|}{\centering \textbf{Overall}}&   \multicolumn{4}{c|}{\centering NLPR} & \multicolumn{4}{c|}{\centering NJU2K} & \multicolumn{4}{c}{\centering STERE}  \\

     & S$\uparrow$
     & F$_\beta$ $\uparrow$ &$E_{\xi}\uparrow$
     & M$\downarrow$ & S$\uparrow$
     & F$_\beta$ $\uparrow$ &$E_{\xi}\uparrow$
     & M$\downarrow$ & S$\uparrow$
     & F$_\beta$$\uparrow$&$E_{\xi}\uparrow$ & M$\downarrow$
     & S$\uparrow$
     & F$_\beta$$\uparrow$ &$E_{\xi}\uparrow$& M$\downarrow$ \\
    \midrule
    RGB
    &[.920]&[.927]&[.951]&[.031]
    &.934&.929&.964&.019
    &.928&.936&.954&.029
    &.931&.937&.960&.026
    \\
    RGB-D (Ours)
    &\colorbox[RGB]{255,230,230}{[\textbf{.930}]}&\colorbox[RGB]{255,230,230}{[\textbf{.938}]}&\colorbox[RGB]{255,230,230}{[\textbf{.960}]}&\colorbox[RGB]{255,230,230}{[\textbf{.027}]}
    &\textbf{.936}&\textbf{.932}&\textbf{.965}&\textbf{.018}
    &\textbf{.932}&\textbf{.941}&\textbf{.957}&\textbf{.028}
    &\textbf{.935}&\textbf{.947}&\textbf{.966}&\textbf{.025}
    \\
   \bottomrule
   \toprule
   \multirow{2}{*}{\centering Methods} &\multicolumn{4}{c|}{\centering SIP}&\multicolumn{4}{c|}{\centering DES} & \multicolumn{4}{c|}{\centering LFSD} & \multicolumn{4}{c}{\centering SSD}\\
     & S$\uparrow$
     & F$_\beta$ $\uparrow$ &$E_{\xi}\uparrow$
     & M$\downarrow$ & S$\uparrow$
     & F$_\beta$ $\uparrow$ &$E_{\xi}\uparrow$
     & M$\downarrow$ & S$\uparrow$
     & F$_\beta$$\uparrow$&$E_{\xi}\uparrow$ & M$\downarrow$& S$\uparrow$
     & F$_\beta$$\uparrow$&$E_{\xi}\uparrow$ & M$\downarrow$\\
    \midrule
    RGB
    &.904&.918&.939&.038
    &.926&.928&.965&.019
    &.882&.885&.913&.060
    &.889&.882&.935&.040
    \\
    RGB-D (Ours)

    &\textbf{.927}&\textbf{.938}&\textbf{.958}&\textbf{.028}
    &\textbf{.930}&\textbf{.933}&\textbf{.967}&\textbf{.017}
    &\textbf{.891}&\textbf{.892}&\textbf{.920}&\textbf{.057}
    &\textbf{.902}&\textbf{.900}&\textbf{.945}&\textbf{.034}\\
     \bottomrule
    \hline
\end{tabular}
}
\label{tab:RGBDModalAblation}
\end{table*}
\subsubsection{Ablation on the number of point prompts from scribbles}
Table \ref{tab:RGBDPromptAblation} gives the ablation study of the number of point prompts from scribbles. We find the model with 10 point prompts from scribbles is better than those with 20 and 30 points. Therefore, we randomly extract 10 points from scribbles as the prompt of SAM.
\begin{table*}[!htp]
\caption{Ablation study on RGB-D SOD datasets about the number of prompt points extracted from scribbles. The best results are bold, and  pink background indicates our overall performance.}
\centering
\resizebox{1\linewidth}{!}
{
   \begin{tabular}{c|cccc|cccc|cccc|cccc}
\hline\toprule
   \multirow{2}{*}{\centering Methods} &\multicolumn{4}{c|}{\centering \textbf{Overall}}&   \multicolumn{4}{c|}{\centering NLPR} & \multicolumn{4}{c|}{\centering NJU2K} & \multicolumn{4}{c}{\centering STERE}  \\
     & S$\uparrow$
     & F$_\beta$ $\uparrow$ &$E_{\xi}\uparrow$
     & M$\downarrow$ & S$\uparrow$
     & F$_\beta$ $\uparrow$ &$E_{\xi}\uparrow$
     & M$\downarrow$ & S$\uparrow$
     & F$_\beta$$\uparrow$&$E_{\xi}\uparrow$ & M$\downarrow$
     & S$\uparrow$
     & F$_\beta$$\uparrow$ &$E_{\xi}\uparrow$& M$\downarrow$ \\
    \midrule
    30 prompt
    &[.917]&[.921]&[.954]&[.032]
    &.928&.925&\textbf{.967}&.020
    &.924&.934&\textbf{.958}&.030
    &.918&.924&.958&.032
    \\
    20 prompt
    &[.921]&[.928]&[.954]&[.029]
    &.935&\textbf{.932}&.966&.019
    &.927&.936&.955&\textbf{.028}
    &.929&.936&.959&\textbf{.025}
    \\
    10 prompt (Ours)
    &\colorbox[RGB]{255,230,230}{[\textbf{.930}]}&\colorbox[RGB]{255,230,230}{[\textbf{.938}]}&\colorbox[RGB]{255,230,230}{[\textbf{.960}]}&\colorbox[RGB]{255,230,230}{[\textbf{.027}]}
    &\textbf{.936}&\textbf{.932}&.965&\textbf{.018}
    &\textbf{.932}&\textbf{.941}&.957&\textbf{.028}
    &\textbf{.935}&\textbf{.947}&\textbf{.966}&\textbf{.025}
    \\

   \bottomrule
   \toprule
   \multirow{2}{*}{\centering Methods} & \multicolumn{4}{c|}{\centering SIP}&\multicolumn{4}{c|}{\centering DES} & \multicolumn{4}{c|}{\centering LFSD} & \multicolumn{4}{c}{\centering SSD}\\
     & S$\uparrow$
     & F$_\beta$ $\uparrow$ &$E_{\xi}\uparrow$
     & M$\downarrow$ & S$\uparrow$
     & F$_\beta$ $\uparrow$ &$E_{\xi}\uparrow$
     & M$\downarrow$ & S$\uparrow$
     & F$_\beta$$\uparrow$&$E_{\xi}\uparrow$ & M$\downarrow$& S$\uparrow$
     & F$_\beta$$\uparrow$&$E_{\xi}\uparrow$ & M$\downarrow$\\
    \midrule
    30 prompt
    &.913&.916&.948&.034
    &.929&.929&\textbf{.970}&\textbf{.017}
    &.877&.885&\textbf{.920}&.062
    &.887&.872&.931&.042\\
    20 prompt
    &.911&.922&.947&.034
    &.923&.928&.968&.019
    &.882&.891&.916&.060
    &.899&.897&\textbf{.946}&.036\\
    10 prompt (Ours)
    &\textbf{.927}&\textbf{.938}&\textbf{.958}&\textbf{.028}
    &\textbf{.930}&\textbf{.933}&.967&\textbf{.017}
    &\textbf{.891}&\textbf{.892}&\textbf{.920}&\textbf{.057}
    &\textbf{.902}&\textbf{.900}&.945&\textbf{.034}\\

     \bottomrule
    \hline
\end{tabular}
}
\label{tab:RGBDPromptAblation}
\end{table*}

\subsubsection{Ablation on the position of modal-aware modulators}
Table \ref{tab:LayerAblation} gives the ablation study of the position of modal-aware modulators. L/3, 2L/3, and 2/L indicate that the  modal-aware modulators are conducted in the last third, two-thirds, and half of the layer, respectively.
As the number of layers increases so do the trainable parameters and FLOPs. However, the overall performances are not incremental. It can be inferred that the  features in the latter half of layers of  SAM transformer encoder are more important when adapting SAM to salient object detection task.
\begin{table*}[!htp]
\caption{Ablation study on RGB-D SOD datasets about the position of  modulators. The best results are bold, and  pink background indicates our overall performance.}
\centering
\resizebox{1\linewidth}{!}
{
   \begin{tabular}{c|c|cccc|cccc|cccc|cccc}
\hline\toprule
   \multirow{2}{*}{\centering Methods} &Trainable &\multicolumn{4}{c|}{\centering \textbf{Overall}}&   \multicolumn{4}{c|}{\centering NLPR} & \multicolumn{4}{c|}{\centering NJU2K} & \multicolumn{4}{c}{\centering STERE}  \\
     &Params(M)
     & S$\uparrow$
     & F$_\beta$ $\uparrow$ &$E_{\xi}\uparrow$
     & M$\downarrow$ & S$\uparrow$
     & F$_\beta$ $\uparrow$ &$E_{\xi}\uparrow$
     & M$\downarrow$ & S$\uparrow$
     & F$_\beta$$\uparrow$&$E_{\xi}\uparrow$ & M$\downarrow$
     & S$\uparrow$
     & F$_\beta$$\uparrow$ &$E_{\xi}\uparrow$& M$\downarrow$ \\
    \midrule
    L/3
    &8.74
    &[.920]&[.930]&[.956]&[.031]
    &.933&.929&.966&.020
    &.928&.934&.956&.030
    &.928&.944&.967&.028
    \\
    2L/3
    &9.74
    &[.924]&[.930]&[.956]&[.029]
    &\textbf{.937}&\textbf{.932}&\textbf{.969}&\textbf{.018}
    &.931&.936&\textbf{.957}&\textbf{.028}
    &\textbf{.935}&.945&\textbf{.968}&\textbf{.024}
    \\
    L/2 (Ours)
    &9.24
    &\colorbox[RGB]{255,230,230}{[\textbf{.930}]}&\colorbox[RGB]{255,230,230}{[\textbf{.938}]}&\colorbox[RGB]{255,230,230}{[\textbf{.960}]}&\colorbox[RGB]{255,230,230}{[\textbf{.027}]}
    &.936&\textbf{.932}&.965&\textbf{.018}
    &\textbf{.932}&\textbf{.941}&\textbf{.957}&\textbf{.028}
    &\textbf{.935}&\textbf{.947}&.966&.025
    \\
   \bottomrule
   \toprule
   \multirow{2}{*}{\centering Methods} &FLOPs&\multicolumn{4}{c|}{\centering SIP}&\multicolumn{4}{c|}{\centering DES} & \multicolumn{4}{c|}{\centering LFSD} & \multicolumn{4}{c}{\centering SSD}\\
   & (G) & S$\uparrow$
     & F$_\beta$ $\uparrow$ &$E_{\xi}\uparrow$
     & M$\downarrow$ & S$\uparrow$
     & F$_\beta$ $\uparrow$ &$E_{\xi}\uparrow$
     & M$\downarrow$ & S$\uparrow$
     & F$_\beta$$\uparrow$&$E_{\xi}\uparrow$ & M$\downarrow$& S$\uparrow$
     & F$_\beta$$\uparrow$&$E_{\xi}\uparrow$ & M$\downarrow$\\
    \midrule
    L/3
    &2633.10
    &.909&.922&.944&.037
    &.928&\textbf{.933}&.965&.018
    &.884&.891&\textbf{.922}&.062
    &.888&.885&\textbf{.946}&.042\\
    2L/3
    &2638.10
    &.909&.917&.943&.037
    &.928&.926&\textbf{.968}&.018
    &.890&\textbf{.894}&.917&\textbf{.056}
    &\textbf{.902}&\textbf{.900}&.944&\textbf{.033}\\
    L/2 (Ours)
    &2635.70
    &\textbf{.927}&\textbf{.938}&\textbf{.958}&\textbf{.028}
    &\textbf{.930}&\textbf{.933}&.967&\textbf{.017}
    &\textbf{.891}&.892&.920&.057
    &\textbf{.902}&\textbf{.900}&.945&.034\\
     \bottomrule
    \hline
\end{tabular}
}
\label{tab:LayerAblation}
\end{table*}

\subsubsection{Ablation on the input resolution}
Table \ref{tab:RGBDInputResolutionAblation} gives the ablation study of the input resolution.
The batchsize can be set as 16 when model adopts 256$\times$256 input resolution in the training stage.
The model has advantage in a few datasets, such as NLPR, DES, and LFSD. However, the overall performance is worse than that of the model with 1024$\times$1024 input resolution.
Meanwhile, the model with 256$\times$256 input resolution is more friendly than that with 1024$\times$1024 in computation cost. FLOPs of the former is 10 times less than that of the latter. Therefore, the model with 256$\times$256 input resolution is also an alternative in limited computation resources.

\begin{table*}[!htp]
\caption{Ablation study on RGB-D SOD datasets about input resolution. The best results are bold, and  pink background indicates our overall performance.}
\centering
\resizebox{1\linewidth}{!}
{
   \begin{tabular}{c|c|cccc|cccc|cccc|cccc}
\hline\toprule
   \multirow{2}{*}{\centering Methods} &Trainable &\multicolumn{4}{c|}{\centering \textbf{Overall}}&   \multicolumn{4}{c|}{\centering NLPR} & \multicolumn{4}{c|}{\centering NJU2K} & \multicolumn{4}{c}{\centering STERE}  \\
     &Params(M)
     & S$\uparrow$
     & F$_\beta$ $\uparrow$ &$E_{\xi}\uparrow$
     & M$\downarrow$ & S$\uparrow$
     & F$_\beta$ $\uparrow$ &$E_{\xi}\uparrow$
     & M$\downarrow$ & S$\uparrow$
     & F$_\beta$$\uparrow$&$E_{\xi}\uparrow$ & M$\downarrow$
     & S$\uparrow$
     & F$_\beta$$\uparrow$ &$E_{\xi}\uparrow$& M$\downarrow$ \\
    \midrule
    256$\times$ 256
    &9.24
    &[.922]&[.928]&[.957]&[.029]
    &.935&\textbf{.933}&\textbf{.967}&\textbf{.018}
    &.925&.929&.955&.030
    &.919&.927&.957&.031\\

    1024$\times$1024 (Ours)
    &9.24
    &\colorbox[RGB]{255,230,230}{[\textbf{.930}]}&\colorbox[RGB]{255,230,230}{[\textbf{.938}]}&\colorbox[RGB]{255,230,230}{[\textbf{.960}]}&\colorbox[RGB]{255,230,230}{[\textbf{.027}]}
    &\textbf{.936}&.932&.965&\textbf{.018}
    &\textbf{.932}&\textbf{.941}&\textbf{.957}&\textbf{.028}
    &\textbf{.935}&\textbf{.947}&\textbf{.966}&\textbf{.025}
    \\
   \bottomrule
   \toprule
   \multirow{2}{*}{\centering Methods} &FLOPs&\multicolumn{4}{c|}{\centering SIP}&\multicolumn{4}{c|}{\centering DES} & \multicolumn{4}{c|}{\centering LFSD} & \multicolumn{4}{c}{\centering SSD}\\
   & (G) & S$\uparrow$
     & F$_\beta$ $\uparrow$ &$E_{\xi}\uparrow$
     & M$\downarrow$ & S$\uparrow$
     & F$_\beta$ $\uparrow$ &$E_{\xi}\uparrow$
     & M$\downarrow$ & S$\uparrow$
     & F$_\beta$$\uparrow$&$E_{\xi}\uparrow$ & M$\downarrow$& S$\uparrow$
     & F$_\beta$$\uparrow$&$E_{\xi}\uparrow$ & M$\downarrow$\\
    \midrule
    256$\times$256
    &245.5
    &.922&.933&.957&.029
    &\textbf{.935}&\textbf{.939}&\textbf{.971}&\textbf{.016}
    &\textbf{.892}&\textbf{.900}&\textbf{.927}&\textbf{.053}
    &.897&.871&.937&.037\\

    1024$\times$1024 (Ours)
    &2635.7
    &\textbf{.927}&\textbf{.938}&\textbf{.958}&\textbf{.028}
    &.930&.933&.967&.017
    &.891&.892&.920&.057
    &\textbf{.902}&\textbf{.900}&\textbf{.945}&\textbf{.034}\\
     \bottomrule
    \hline
\end{tabular}
}
\label{tab:RGBDInputResolutionAblation}
\end{table*}

\section{Conclusions}
In the work, \textit{SSFam} is proposed based on the large model SAM for the scribble supervised salient object detection task in unimodal, bimodal, and trimodal images. The modal-aware modulators are responsible for attaining modal-specific information, adapting to the input with combinations of different modalities. The siamese decoder transfers the knowledge from prompt branch to no prompt branch, reducing the gap between the prompt training and no prompt testing. Experiments delivers the outstanding performance across RGB, RGB-D, RGB-T, and V-D-T SOD tasks.
Due to the high computation cost of \textit{SSFam} based on SAM, the real-time deployment of model is challenging, which is  reserved  for future research.


\end{document}